\documentclass[review]{elsarticle}

\usepackage{lineno,hyperref}

\journal{Expert Systems with Applications}
\graphicspath{ {images/} }
\usepackage{nccmath}
\usepackage{comment}
\usepackage{adjustbox}
\usepackage{amsmath}
\usepackage{mathtools}
\usepackage{booktabs}
\usepackage{multirow}
\usepackage{tabularx}
\usepackage{amsfonts}
\DeclareMathOperator*{\argmin}{min}

\usepackage{pifont}
\newcommand{\cmark}{\ding{51}}
\newcommand{\xmark}{\ding{55}}
\newcommand\tabitem{\makebox[1em][r]{\textbullet~}}
\usepackage{array}
\newcolumntype{P}[1]{>{\raggedright\arraybackslash}p{#1}}
\usepackage{caption} 
\begin{document}

\begin{frontmatter}

\title{DarkDeblur: Learning single-shot image deblurring in low-light condition}
\author[address1]{S M A Sharif}
\ead{sma.sharif.cse@ulab.edu.bd}

\author[address2]{Rizwan Ali Naqvi \corref{cor1}}
\cortext[cor1]{Corresponding author}
\ead{rizwanali@sejong.ac.kr}

\author[address3]{Farman Ali}
\ead{farmankanju@sejong.ac.kr}

\author[address1]{Mithun Biswas}
\ead{mithun.bishwash.cse@ulab.edu.bd }

\address[address1]{Rigel-IT, Banasree, Dhaka-1219, Bangladesh}
\address[address2]{Department of Unmanned Vehicle Engineering, Sejong University, South Korea}
\address[address3]{Department of Software, Sejong University, South Korea}

\fntext[note]{Code and dataset available at: https://github.com/sharif-apu/DarkDeblur}

\begin{abstract}
Single-shot image deblurring in a low-light condition is known to be a profoundly challenging image translation task. This study tackles the limitations of the low-light image deblurring with a learning-based approach and proposes a novel deep network named as DarkDeblurNet. The proposed DarkDeblur- Net comprises a dense-attention block and a contextual gating mechanism in a feature pyramid structure to leverage content awareness. The model additionally incorporates a multi-term objective function to perceive a plausible perceptual image quality while performing image deblurring in the low-light settings. The practicability of the proposed model has been verified by fusing it in numerous computer vision applications. Apart from that, this study introduces a benchmark dataset collected with actual hardware to assess the low-light image deblurring methods in a real-world setup. The experimental results illustrate that the proposed method can outperform the state-of-the-art methods in both synthesized and real-world data for single-shot image deblurring, even in challenging lighting environments.
\end{abstract}

\end{frontmatter}

\section{Introduction}
Digital cameras have illustrated a promising performance gain over recent years. Despite evolving in both hardware and image processing aspects, image sensors still suffer from quantum inefficiency in low-light conditions \citep{liu2014effect}. Due to these shortcomings, digital cameras compel to employ long-exposure settings in inferior lighting environments. Therefore, it is frequent to encounter undesired blur artifacts while taking image shots with a hand-held setup expressly in stochastic conditions 
\citep{hu2014deblurring}. Most notably, such blind motion blurs are inevitable and substantially deteriorate the perceptual image quality \citep{schuler2015learning,  shao2020deblurgan+}.       

Perceptual quality enhancement from degraded blurry images refers to a deconvolution operation, where the intended image comprises a blur kernel with additive sensor noises \citep{schuler2015learning}. A substantial amount of noise factors can make the restoration process considerably complicated. Notably, in low-light conditions, image sensors capture the sensor noises most \citep{chatterjee2011noise}. Also, a single-shot image deblurring process does not incorporate reference information of the motion trajectory from neighbor frames \citep{wang2012kernel}. Therefore, a single-shot latent image restoration process in a low-light environment is far more challenging than a typical well-lit image deblurring process. 

\begin{figure}[ht]
\centering
\includegraphics[width=\textwidth,keepaspectratio]{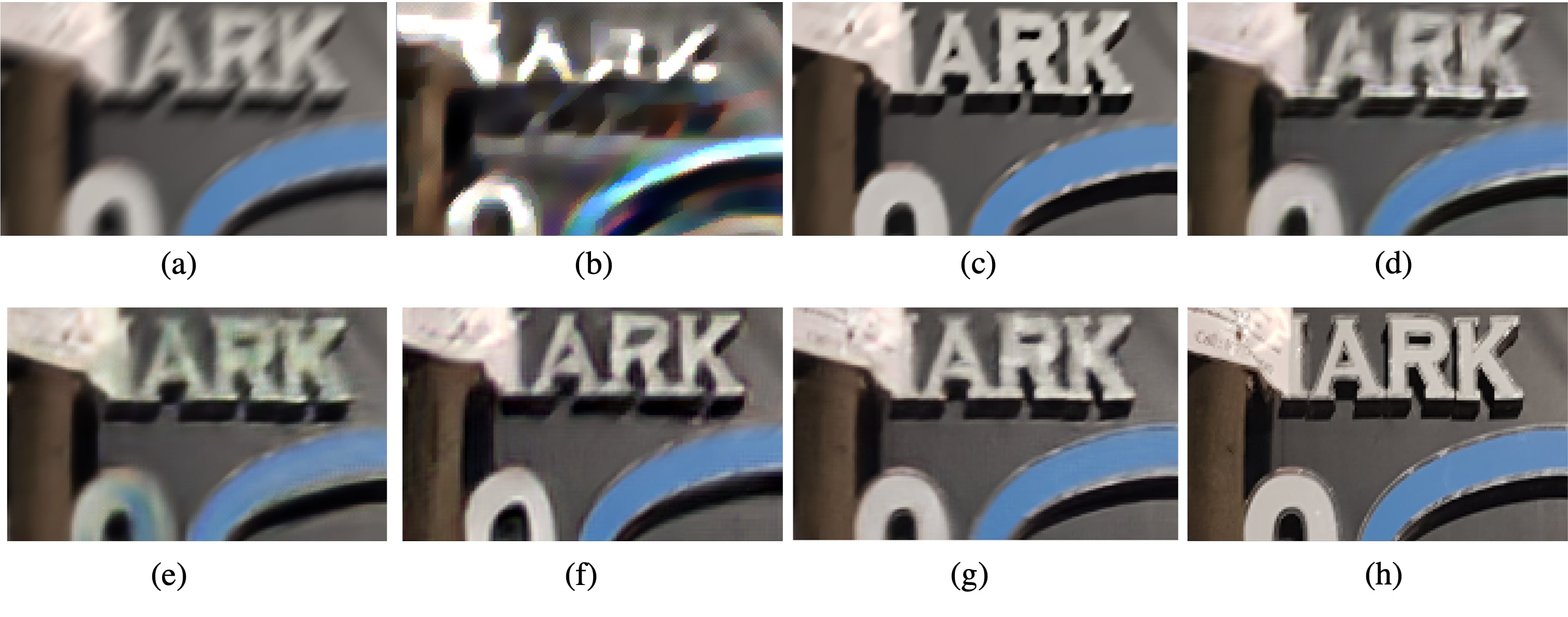}
\caption{ Performance of the SOTA single-shot deblurring methods and proposed DarkDeblurNet on a real-world blurry image. It is visible that existing methods likely to produce visually disturbing artifacts while removing blurs from an image capture in low-light. (a) Blurry input image (PSNR: 23.7). (b) Result of LightStreaks (PSNR: 15.73 ). (c) Result of DarkChannel\citep{pan2016blind} (PSNR:23.44). (d) Result of DeepDeblur \citep{nah2017deep} (PSNR: 25.89). (e) Result of SRN \citep{tao2018scale} (PSNR: 25.77). (f) Result of DeblurGANv2 \citep{kupyn2019deblurgan} (PSNR: 24.44). (g) \textbf{DarkDeblurNet (PSNR:27.99)}. (h) Reference sharp image.}
\label{intro}
\end{figure}

In recent years, learning-based image deblurring methods \citep{kupyn2019deblurgan, nah2017deep, tao2018scale, sun2015learning, schuler2015learning, sim2019deep} have drawn significant momentum in the single-shot image deblurring domain. However, most recent studies focused on the blind deconvolution process without explicitly considering the lighting conditions. Arguably, the lightning condition has a direct impact on the blur removal process.  Nevertheless, to study the feasibility of the state-of-the-art (SOTA) deblurring method in low-light conditions, an initial experiment has been conducted. It is perceptible that even the SOTA single-shot image deblurring methods illustrate the deficiencies and are susceptible to producing visually disturbing artifacts while removing blurs in low-light conditions (please refer to Figure \ref{intro}). 

Despite the unsatisfactory performance of the SOTA methods, the single-shot image deblurring in low-light conditions has numerous promises in computer vision, robot vision, autonomous driving, surveillance solutions, etc. For instance, in recent years, night photography with single-shot hand-held cameras like the smartphone has gained noteworthy interest among end-user \citep{a2021beyond, sharif2021sagan}. Regrettably, motion blurs are inclined to appear most in night shots and deliver an inadmissible photography experience. Contrarily, single-shot image deblurring can enhance the perceptual quality of such vulnerable image samples. Apart from that, low-light image deblurring has numerous real-world applications in autonomous driving and surveillance solutions. Also, single-shot image deblurring can accelerate the performance of computer vision applications (e.g., segmentation, detection, recognition, classification, etc.). The widespread applicability of such image enhancement inspired this study to tackle the challenges of single-shot low-light image deblurring.

This study proposes a novel content-aware learning approach to address the deficiencies of low-light image deblurring. The proposed deep model appropriates the channel attention \citep{hu2018squeeze} in a multi-level feature pyramid structure \citep{lin2017feature, kirillov2019panoptic, lai2017deep} for global image correction. Also, the proposed model utilizes a contextual gating mechanism \citep{liu2018deep, yu2019free} to leverage spatial enhancement in a residual manner \citep{he2016deep, sha2019fast}. The proposed model has optimized with multi-term loss function combining the reconstruction, structure, perceptual features, and adversarial guidance to perceive visually pleasurable images. Apart from that, a blur-sharp image dataset has been developed to assess the performance of deblurring methods in low-light conditions. It is worth noting that the image pairs of the proposed dataset were collected with actual hardware rather than simulating blur artifacts on synthesized data. This study denoted the proposed deep model and real-world benchmark dataset as the "DarkDeblurNet" and "DarkShake" in the rest sections. The contributions of this study are as follows:

\begin{itemize}
  \item DarkDeblurNet: A content-aware deep network specialized in single-shot image deblurring in low-lit conditions is proposed. The model utilizes a novel dense-attention and contextual gating mechanisms in a feature pyramid structure to correct the global and spatial information of a latent sharp image. The proposed dense-attention and contextual gate strive to mitigate the visual artifacts, which commonly appear in the low-light image deblurring.

  \item Multi-term losses: A novel objective function is introduced, which combines multiple losses such as reconstruction loss, structure loss, perceptual feature loss, and an adversarial loss. It allows this study to obtain visually plausible images in challenging conditions.

  \item DarkShake dataset: A blur-sharp image dataset has collected by employing actual hardware. It helps to study the feasibility of the proposed method with real-world blurry images. Also, it intends to serve as a benchmark dataset to assess the performance of the single-shot image deblurring methods in a real-world setup, particularly in low-light conditions.

  \item Dense experiments: The sophisticated experiments illustrate that the proposed method can outperform the SOTA methods in both synthesized and real data. Also,  a series of experiments unveil the practicability of the proposed components as well as the proposed method in numerous computer vision applications. 
\end{itemize}

The rest of this paper is structured into five additional sections. Section \ref{relatedWork}  reviews the related works. Section \ref{method} describes the proposed method. Section \ref{results} illustrates the experimental results. Section \ref{discussion} discusses the limitations as well as future scopes. Finally, section \ref{conclusion} concludes this work.

\section{Proposed Method}
\label{method}
A novel learning-based low-light image deblurring method has been proposed in this study. This section details the designing method of the deep model,  multi-term optimization scheme, dataset preparation, and implementation process consecutively.  
\subsection{Network Design}
Typically, deblurring an image through the deconvolution process is considered a variant of image-to-image translation. Hence, the proposed DarkDeblurNet set the aim of deblurring in low-light conditions as $\mathrm{F}: I_B \to  I_D$. Where mapping function ($\mathrm{F}$) learns to generate a sharper image ($I_D$) from a blurry input ($I_B$) as $I_D \in [0,1]^{H \times W \times 3}$. $H$ and $W$ represent the height and width of the input as well as output images. 

\begin{figure}[!htb]
\centering
\includegraphics[width=\textwidth,keepaspectratio]{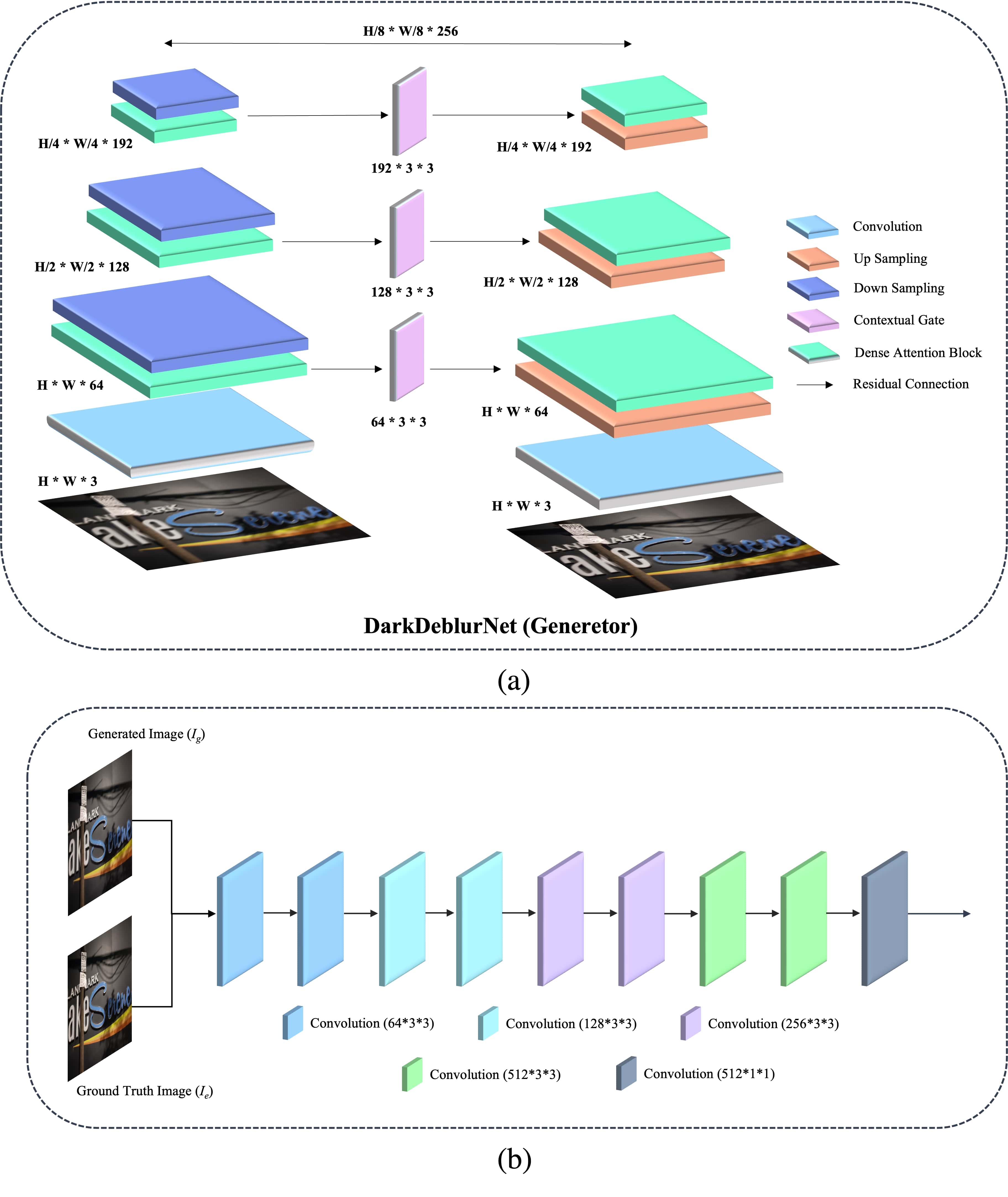}
\caption{ The overview of the proposed DarkDeblurNet. The proposed network incorporates a novel dense-attention block and contextual gating mechanism in a feature pyramid structure. Also, it follows the principle of generative adversarial networks. (a) The architecture of the generator. (b) The architecture of the discriminator.}
\label{gen}
\end{figure}

As Figure \ref{gen} illustrates, the proposed DarkDeblurNet incorporates the concept of GAN. Here, the generator of the proposed DarkDeblurNet design is such that it can utilize the advantages of a feature pyramid structure with a novel dense-attention block. Also, the features learned at different feature-level have propagated with a contextual gating mechanism to leverage spatial awareness.

\begin{figure}[ht!]
\centering
\includegraphics[width=8cm,keepaspectratio]{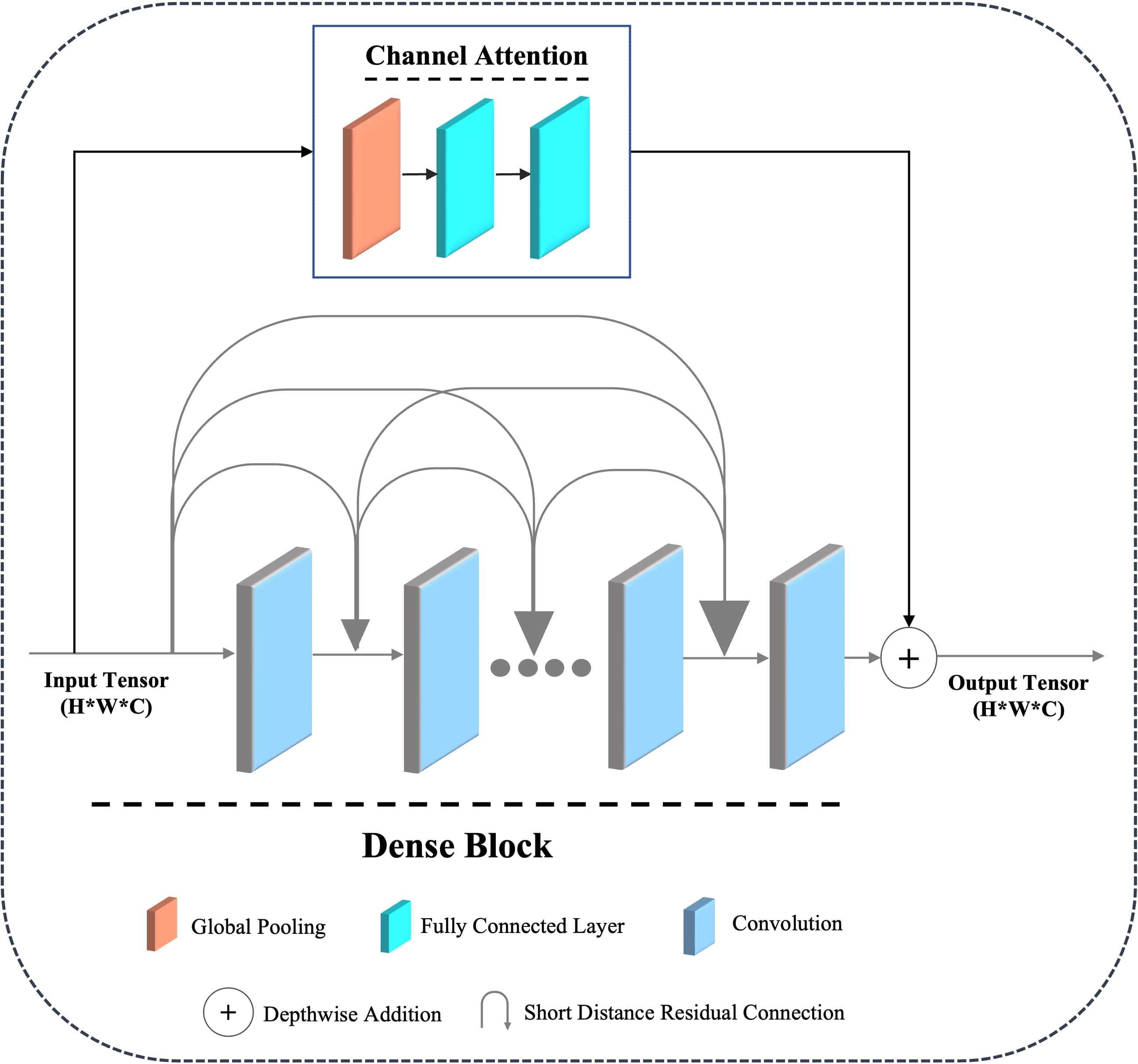}
\caption{Overview of the proposed dense-attention block. The dense-attention block combines a residual dense block and channel attention mechanism. It aims to capture the global feature interdependencies in different feature levels.}
\label{denseAttention}
\end{figure}

Dense-attention block: Figure \ref{denseAttention} depicts the overview of the proposed dense-attention block. The key idea of a dense-attention block is to go deeper and wider along with learning global feature interdependencies. The proposed dense-attention block developed by taking inspiration from residual-dense block \citep{zhang2018residual, wang2018esrgan} and squeeze-extraction networks \citep{hu2018squeeze}, which were proposed in recent studies. 

Typically, a dense block connect all the previous layers as
\begin{equation}
X_{l} = H_l([X_0,X_1,...,X_{l-1}])
\end{equation}

Where $[X_0,X_1,...,X_{l-1}]$ refers to the concatenated
feature maps obtained through the $l=5$ number of convolutional layers \citep{huang2017densely}. This study set the number of convolutional layers ($l$) as $l=5$. All convolution layers have comprised the kernel = $3 \times 3$, stride = 1, padding = 1, and activated with LeakyReLU function.

On the other hand, global feature interdependencies can perceive by applying a global average pooling \citep{hu2018squeeze, dai2019second}. Where channel-wise squeezed descriptors $Z \in \mathbb{R}^C$ of an input feature map can be calculated as:
 
\begin{equation}
Z_{c} = \frac{1}{H \times W}\sum_{i=1}^{H}\sum_{j=1}^{W}U_{c}(i, j) 
\end{equation}
 
Here, $Z_{c}$,  $H \times W$, and $U$  present the global average pooling, spatial dimension, and feature map.

To pursue aggregated global dependencies, a gating mechanism has applied as follows :
 
\begin{equation}
W = \sigma(W_{E}(\delta(W_{S}(Z))))
\end{equation}

Here, $\sigma$ and $\delta$ denote sigmoid and ReLU activation functions applied after $W_{E}$ and $W_{S}$ convolutional operations.

The final channel attention map has achieved by rescaling the feature map as follows:
 
\begin{equation}
\hat{S}_{c} = W_{c}.S_{c}
\end{equation}

Here, $W_{c}$ and $S_{c}$ denote the scaling factor and feature map. Typically, squeeze and extraction-based channel attention $\hat{S}_{c}$ are calculated over the output of convolutional blocks (i.e., residual block, dense block, etc.) \citep{hu2018squeeze}. However, this study proposes to calculate the squeeze and extraction descriptors from the input of the convolutional block and propagate it as a residual connection to perceive long-distance channel-wise attention.

The final output of the dense-attention block has obtained as follows:
 
\begin{equation}
D_{a} = X_{l} + \hat{S}_{c}
\end{equation}

\begin{figure}[ht]
\centering
\includegraphics[width=8cm,keepaspectratio]{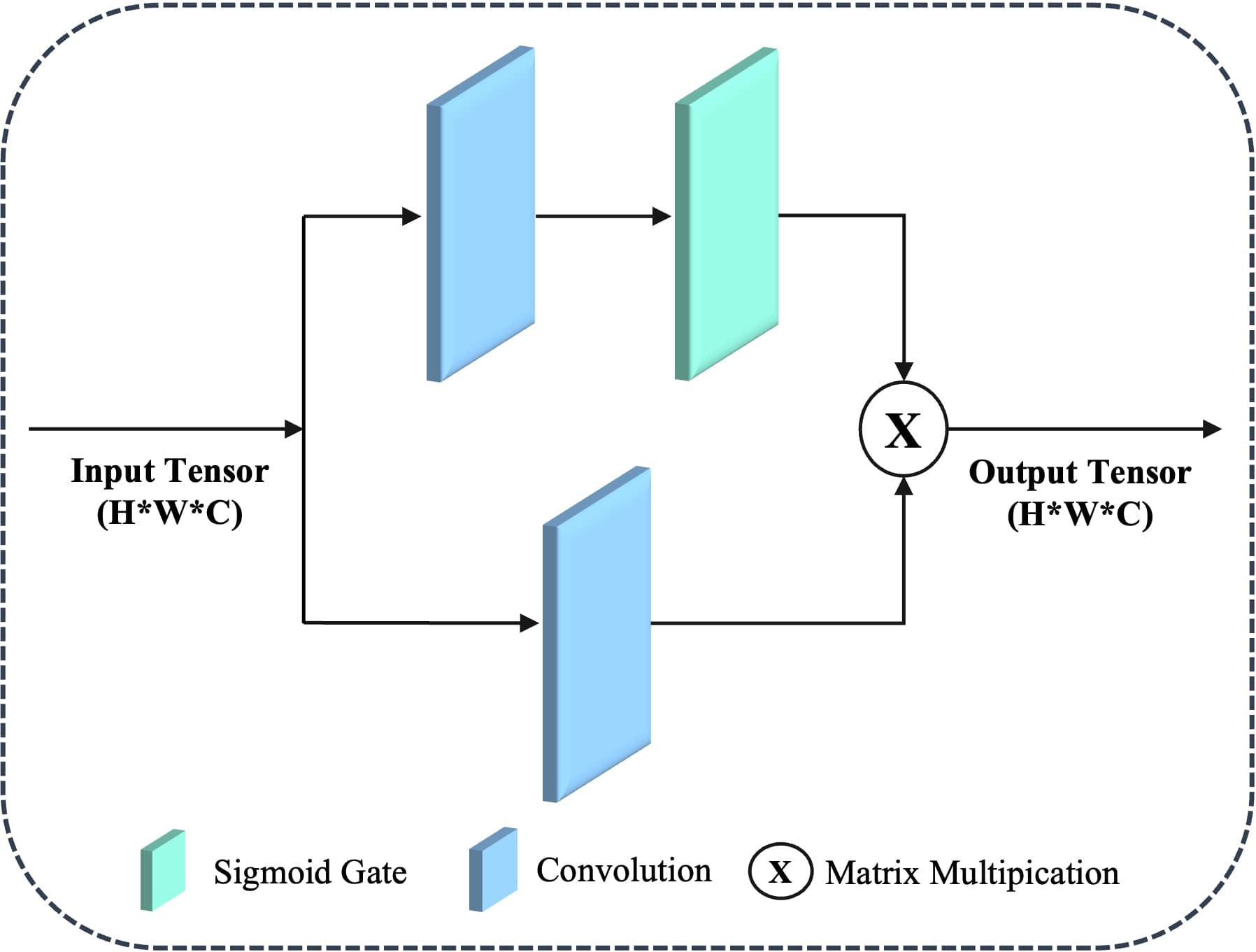}
\caption{ Overview of the contextual gate. The contextual-gate aims to propagate the spatial dependencies between different feature levels.  
 }
\label{CG}
\end{figure}

Contextual gate: 
Figure \ref{CG} illustrates the overview of the context gate. Here, the contextual gate aims to propagate the important feature only \citep{yu2019free}. Despite a typical residual connection \citep{szegedy2016inception}, the context gate does not pass trivial features from the lower level. The context gate has obtained as follows: obtained as follows:

\begin{equation}
G_{m,n} = \sum_{m=1}^{H}\sum_{n=1}^{W} W_{g}.I 
\end{equation}

\begin{equation}
F_{m,n} = \sum_{m=1}^{H}\sum_{n=1}^{W} W_{f}.I 
\end{equation}

\begin{equation}
O_{m,n} = \phi(G_{m,n})	\odot \delta(F_{m,n})
\end{equation}
 
Here, $ \phi $ and $\delta$ present the LeakyReLU and sigmoid activations. $W_{g}$ and $W_{f}$ represent convolutional operations.

Level Transition. The proposed DarkDeblurNet traverse different features dimension (i.e., upscaling or downscaling) with convolution operation. Here, the downsampling operation has been obtained as follows:
 
\begin{equation}
F_{\downarrow} = H_{\downarrow} (X_{0})
\end{equation}
 
Here, $H_{\downarrow}$ represents the down sampling through the convolutional operation comprises a kernel = $3 \times 3$, stride = 2, padding = 1.

Inversely, the upscaling has obtained as follows:
 
\begin{equation}
F_{\uparrow} = H_{\uparrow} (X_{0})
\end{equation}

Here, $H_{\uparrow}$ represents the pixel shuffle convolution operation \citep{aitken2017checkerboard}, activated with a PReLU function. The pixel shuffle convolution aims to avoid checkerboard artifacts \citep{aitken2017checkerboard, shao2020deblurgan+}.

Conditional Discriminator:
The proposed DarkDeblurNet appropriates the concept of adversarial guidance. This study adopted a well-established variant of GAN known as conditional GAN (cGAN) \citep{mirza2014conditional, liu2020importance}. Therefore, the goal of the discriminator has been set to maximize $\mathbb{E}_{X,Y} \big[\log D\big(X,Y\big) \big]$. The overall discriminator architecture can be considered a stacked Convolutional Neural Network (CNN). Where all layers before the output layer include convolutional kernel =$3 \times 3$, followed by batch normalization, and activated with a swish function. The feature depth of the discriminator started from 64 channels. Every $(2n-1)^{th}$ layer increases the feature depth and reduces the spatial dimension by a factor of 2. Here, the spatial dimension has been reduced by applying a stride = 2. The output from the final layer obtains with a convolution operation, where the layer comprises a  kernel = $1 \times 1$ and is activated by a sigmoid function.

\subsection{Objective Function}
The proposed DarkDeblurNet learns to deblur with a sophisticated mapping function ($\mathrm{F}$) with parameterized weights $W$. Given the training set $\{ I_B^t, I_S^t \}_{t=1}^P$ consisting of $P$ image pairs, the training process aims to minimize the objective function describes as follows:

\begin{equation}
 W^\ast = \arg{\argmin_W}\frac{1}{P}\sum_{t=1}^{P}\mathcal{L}_{\mathit{T}}(\mathrm{F}(I_B^t), I_S^t)
 \label{fLoss}
\end{equation}

Here, $\mathcal{L}_{\mathit{T}}$ denotes the proposed multi-term loss for single-shot image deblurring in low-light conditions. The goal of this multi-term loss is to improve the perceptual quality (i.e., details, texture, color, structure, etc.) of a given blurry image.

Reconstruction loss:
A pixel-wise loss is adopted to perceive a coarse to refine reconstruction. Typically, an L1 or an L2 distance is used as a pixel-wise loss function. However, among these two, L2-loss is directly related to the PSNR and tends to produce smoother images \citep{schwartz2018deepisp}. Therefore, an L1 objective function has been considered as reconstruction loss in this study, which was calculated in the training phase as follows:

\begin{equation}
 \mathcal{L}_{\mathit{R}} = \parallel I_S-I_D \parallel_1
\end{equation}

Here, $I_D$ and $I_S$ present the output obtain through $\mathrm{F}(I_B)$ and reference sharp image.

Structure loss: One of the drawbacks of the deblurring method is to produce structural distortion while deblurring images. To address this limitation, this study proposes to utilize a structural similarity loss as one of the objective functions. The previous studies reported that an SSIM loss works well with an L1-loss to improve structural deficiencies \citep{schwartz2018deepisp, zhao2016loss}. Thus, SSIM loss has been utilized as a structural loss and calculated as follows:

\begin{equation}
 \mathcal{L}_{\mathit{S}} = SSIM \Big( I_S, I_D \Big)
\end{equation}

Here, a multi-scale variant of SSIM-loss has been used.

Perceptual Feature loss:
The perceptual loss was introduced based on the activation maps produced by the ReLU layers of the pre-trained VGG-19 network \citep{johnson2016perceptual, ledig2017photo, wang2018esrgan, gai2019new}. Instead of measuring the per-pixel difference between two images, this loss prompts images to have similar feature representation that comprises to obtain a similar perceptual quality. It is worth noting, the perceptual loss works best with the L1 norm perceived from the top-layer \citep{wang2018esrgan, ignatov2017dslr}. Otherwise, the perceptual feature loss can produce inconsistent brightness and visual artifacts. However, the previous study on deblurring \citep{kupyn2019deblurgan, kupyn2018deblurgan} that utilizes the perceptual loss does not incorporate the basic principle. Therefore, this study defined perceptual loss as perceptual feature loss and formulated as follows:

\begin{equation}
 \mathcal{L}_{\mathit{F}} = 
 \frac{1}{H_j \times W_j \times C_j}\parallel \psi_t(I_S) - \psi_t(I_D) \parallel_1
\end{equation}

Here, $\psi$ and $j$ denote the pre-trained VGG network and $j^{th}$ layer.

Adversarial loss:
GANs illustrated the superior performance in producing sharper and realistic images by observing generated and reference images \citep{kupyn2018deblurgan}. They are also capable of recovering textures from smoother inputs \citep{ignatov2017dslr, wang2018esrgan}. The cGAN loss utilized in this study incorporates to minimize the cross-entropy loss function as follows:

\begin{equation}
 \mathcal{L}_{\mathit{G}}= - \sum_{t} \log D(I_D, I_S)
\end{equation}
 
Here, $D$ denotes the conditional discriminator used as a global critic.

Multi-term loss:
The  multi-term objective  ($\mathcal{L}_{\mathit{T}}$) has obtained as follows:
 
\begin{equation}
 \mathcal{L}_{\mathit{T}}= \mathcal{L}_{\mathit{R}} + \mathcal{L}_{\mathit{S}} + \lambda_{F}. \mathcal{L}_{\mathit{F}} +  \lambda_{G}.\mathcal{L}_{\mathit{B}}
\end{equation}
 
Here, $\lambda_{F}$ and $\lambda_{G}$ present the loss regulators, which set as  $\lambda_{F}$ = 1e-2 and $\lambda_{G}$ = 1e-4 to stabilize the adversarial training.

\subsection{Dataset Preparation}
Dataset preparation is a crucial part of the learning and evaluation process. It is harder to obtain a substantial amount of images \citep{sim2019deep, kupyn2018deblurgan}, which pair into blurry and sharp frames. This study addressed the data limitation by leveraging synthesizing and real-world data samples.

\subsubsection{Training Dataset}
\begin{figure}[!htb]
\centering
\includegraphics[width=\textwidth,keepaspectratio]{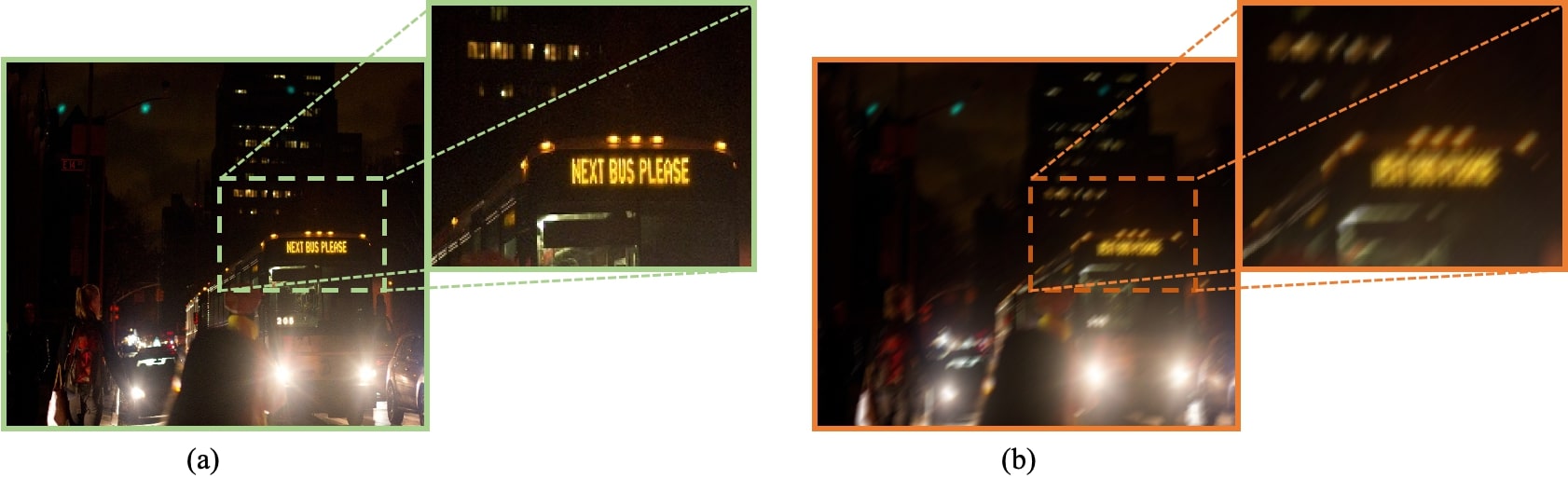}
\caption{ Example of a blur-sharp image pair obtained from the ExDark dataset. (a) Reference (sharp) image. (b) Blurry (simulated) image. }
\label{ExDarkExm}
\end{figure}

The existing dataset available for image deblurring doesn't specialize in low-light conditions. Hence, a benchmark low-light dataset known as ExDark \citep{Exdark} is employed to study the proposed method. The ExDark dataset contains 7,363 low-light images with a variety of different objects and scenes. This study divides the ExDark dataset for training and evaluation sets. The training and testing set comprises 6800 and 563 images, which are selected arbitrarily. It is worth noting, the images obtained from the ExDark dataset do not include the blur-sharp image pairs as required for training and evaluation. Therefore, a Markovian motion trajectory generation process \citep{kupyn2019deblurgan, kupyn2018deblurgan} with random blur-kernel has applied to the ExDark images to obtain the motion blurs. A total of 559,290 non-overlapping image patches has extracted for training purposes. Each blur-sharp image pair of training set comprise an image dimension of $128 \times 128 \times 3$. Also, during training phase a random noise factor ($\mathcal{N}\mathbf{I_N}|\sigma)$ has been applied to the training samples to replicate the sensor noise. Here, $\sigma$ represents the standard deviation of the noise distribution, which is generated by $\mathcal{N}(\cdot)$. Figure \ref{ExDarkExm} illustrates the image pair of sharp-blur images.

Apart from learning from the synthesized dataset, the proposed method has also been studied with existing dynamic deblurring datasets (i.e., GoPro \citep{nah2017deep} and REDS \citep{nah2017deep}). It is worth noting that the proposed study focuses on challenging single-shot image deblurring. Hence, every second frame of both datasets has been leveraged to extract  225,100 non-overlapping images patches for training purposes.

\subsubsection{DarkShake Dataset}
Due to the data limitation, numerous studies used synthesized data for evaluation purposes. However, the synthesized image samples with simulated motion blur artifacts may not reflect the real-world blurry images. Therefore, this study incorporates a novel blur-sharp image pair collection method, which uses the actual hardware for collecting the blur-sharp image pairs.

\begin{figure}[!htb]
\centering
\includegraphics[width=10cm,keepaspectratio]{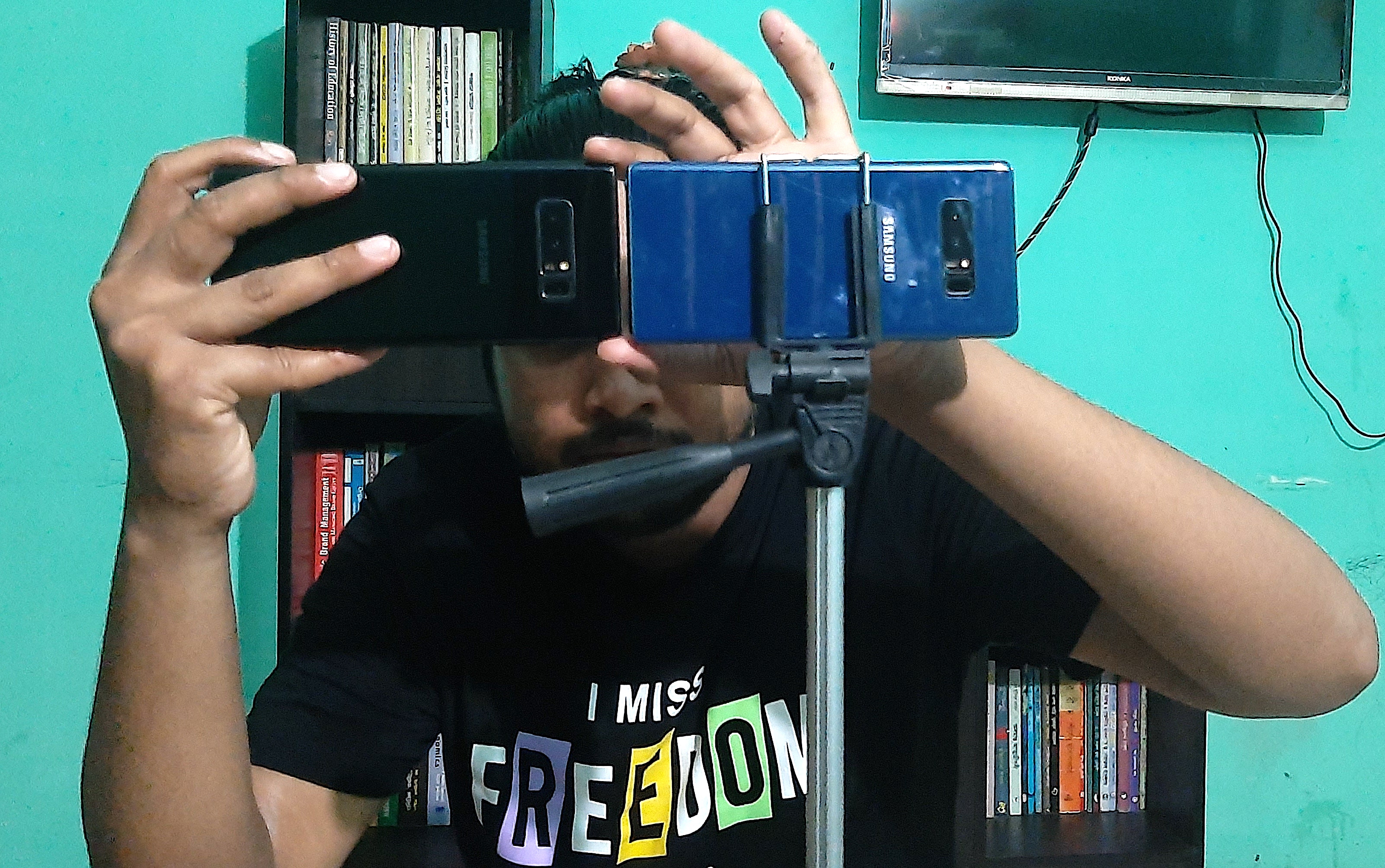}
\caption{Hardware setup was used for capturing the real-world blur-sharp image pairs. Here, one device was mounted into a tripod, where another device was held with bare hands. Both devices were configured evenly to capture the image pairs.}
\label{hardwareSetup}
\end{figure}

Hardware Setup: As illustrated in Figure \ref{hardwareSetup}, a setup comprised of two point-shot cameras (i.e., Galaxy Note 8, Galaxy Note 10+, Galaxy A53, etc., smartphones) is strived to capture the real-world blur-sharp image pairs. Here, one device is mounted into a tripod and aimed to capture the stabilized unblurry images. Contrarily, another camera is held with a bare hand to capture real-world blurry scenes. Both cameras were evenly calibrated and focused on the same surface and taken with long-exposure settings. The shutter speed of the cameras was set under 1 second and the ISO setting was fixed at 800. Both cameras were triggered simultaneously with a wireless command. Overall, we captured 100 image pairs in the lighting condition under 200 lux.

\begin{figure}[!htb]
\centering
\includegraphics[width=\textwidth,keepaspectratio]{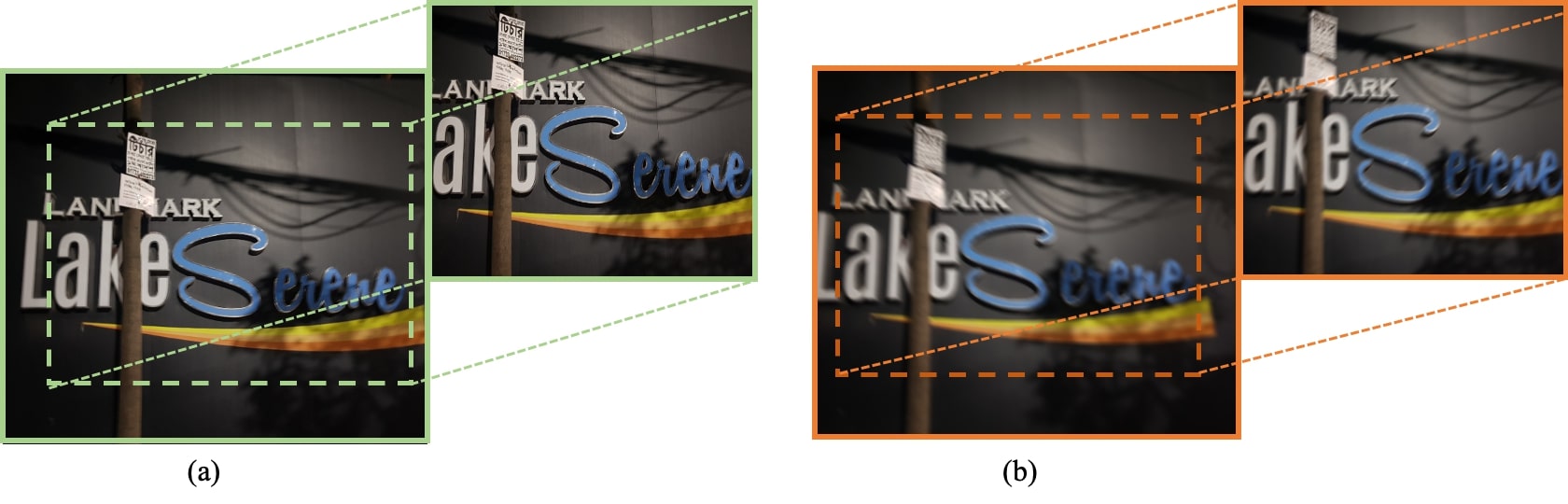}
\caption{ Example of a blur-sharp image pair collected with actual hardware. Both images are aligned for evaluating the deep models. (a) Reference image. (b) Blurry image. }
\label{DarkShakeExm}
\end{figure}

Matching Algorithms:
Due to sifting between the tripod and hand-held setup, the cameras can cover a  different field of view (FOV). Although the capture images are not significantly unaligned, nonetheless, the image pair need to be restricted to a similar FOV \citep{cortes2015interactive}. Therefore, a matching strategy is applied to align the images with an additional non-linear transformation. The SIFT keypoints \citep{lowe2004distinctive} of captured images are calculated, which again used to find a homography matrix using the RANSAC algorithm \citep{vedaldi2010vlfeat}. Later, the match portions were cropped to make the final blur-sharp image pair. Figure \ref{DarkShakeExm} illustrates an example blur-sharp image pair from the proposed DarkShake dataset. 

\subsection{Implementation Detail}
The DarkDeblurNet has been implemented with the PyTorch framework \citep{pytorch}. The proposed generator leverages four consecutive feature levels (i.e., 64, 128, 192, 256, etc.) in the multi-level feature pyramid structure. Apart from that, the contextual gates of the proposed generator utilize filter sizes of 64, 128, and 192 to match the depth of feature levels. Figure 2 depicts the network details.

The Adam optimizer \citep{kingma2014adam}  has been utilized to optimize the generator and discriminator of the proposed DarkDeblurNet. The hyperparameters of both network networks are tuned as $\beta_1 = 0.9$, $\beta_2 = 0.99$, and learning rate = 1e-4.  The DarkDeblurNet was trained with image patches of $128 \times 128 \times 3$ for 100,000 steps with a constant batch size of 16. All experiments were conducted on a machine running on Ubuntu 20.04. The hardware comprises an AMD Ryzen 3200G central processing unit (CPU) clocked at 3.6 GHz, a random-access memory of 16 GB, and An Nvidia Geforce GTX 1060 (6GB) graphical processing unit (GPU).

\section{Experiments and Results}
\label{results}
 This section detail the results and applications of the proposed method with distinct experiments.

\subsection{Comparison with Existing Methods}
The proposed method is trained with synthesized and dynamic deblurring datasets. Nevertheless, the feasibility of the proposed DarkDeblurNet is verified with synthesized and real-world blurry images. The results obtained from the proposed method are compared with the SOTA single-shot learning-based image deblurring methods. The following SOTA methods were selected for the comparison; 1) DeepDeblur, 2) Scale-Recurrent-Network (SRN), and 3) DeblurGANv2. It is worth noting, none of the existing methods is developed to handle the single-shot image deblurring in low-light conditions. For a fair comparison, all target models are retrained with low-light blurry images and suggested hyper-parameters. Apart from that, we also studied the feasibility of non-learning deblurring methods with diverse data samples. Following deblurring methods have been incorporated throughout the proposed study: 1) LightStreaks and 2) Dark Channel Prior. 

\subsubsection{Low-light Deblurring (Synthesize Data)}

Table \ref{EDTAB} depicts the performance of the evaluated methods. The results were obtained with different synthesized datasets: 1) ExDark, 2) Lai Dataset, and 3) Kohler Dataset. It is worth noting, handheld image deblurring is considered as a special case of dynamic deblur. However, to study further,  night shot images from Lai and Kohler datasets have been included. The performance of deblurring methods has been summarized with three evaluation metrics: PSNR, SSIM, and  DeltaE\citep{gomez2016comparison}. The DeltaE metric intends to evaluate the color and brightness consistency obtained through the deblurring methods.

\begin{table}[!htb]
\centering
\caption{Quantitative comparison between SOTA methods and proposed DarkDeblurNet on synthesized datasets. In all evaluation metrics, the DarkDeblurNet outperformed the exiting methods.}


\scalebox{.45}{\begin{tabular}{lllllllllllll}
\hline
\multirow{2}{*}{\textbf{Method}} & \multicolumn{3}{l}{\textbf{Exdark Dataset}}                                                 & \multicolumn{3}{l}{\textbf{Lai Dataset}}                                                    & \multicolumn{3}{l}{\textbf{Kohler Dataset}}                                                 & \multicolumn{3}{l}{\textbf{Average}}                                                                        \\ \cline{2-13} 
                                 & \multicolumn{1}{l}{\textbf{PSNR $\uparrow$}}  & \multicolumn{1}{l}{\textbf{SSIM $\uparrow$}}   & \textbf{DeltaE $\downarrow$} & \multicolumn{1}{l}{\textbf{PSNR $\uparrow$}}  & \multicolumn{1}{l}{\textbf{SSIM $\uparrow$}}   & \textbf{DeltaE $\downarrow$} & \multicolumn{1}{l}{\textbf{PSNR $\uparrow$}}  & \multicolumn{1}{l}{\textbf{SSIM $\uparrow$}}   & \textbf{DeltaE $\downarrow$} & \multicolumn{1}{l}{\textbf{PSNR $\uparrow$}}        & \multicolumn{1}{l}{\textbf{SSIM $\uparrow$}}        & \textbf{DeltaE $\downarrow$}      \\ \hline
LightStreaks \citep{hu2014deblurring}                     & \multicolumn{1}{l}{20.06}          & \multicolumn{1}{l}{0.7041}          & 6.69            & \multicolumn{1}{l}{17.90}           & \multicolumn{1}{l}{0.6825}          & 8.39            & \multicolumn{1}{l}{19.45}          & \multicolumn{1}{l}{0.6700}            & 8.49            & \multicolumn{1}{l}{19.14}          & \multicolumn{1}{l}{0.6855}          & 7.86         \\ 
DarkChannel\citep{pan2016blind}               & \multicolumn{1}{l}{28.16}          & \multicolumn{1}{l}{0.8944}          & 2.78            & \multicolumn{1}{l}{19.68}          & \multicolumn{1}{l}{0.7405}         & 5.95          & \multicolumn{1}{l}{22.65}          & \multicolumn{1}{l}{0.7855}          & 5.03            & \multicolumn{1}{l}{23.50}          & \multicolumn{1}{l}{0.8068}              & 4.59          \\ 
DeepDeblur \citep{nah2017deep} (f) Result of                      & \multicolumn{1}{l}{31.40}           & \multicolumn{1}{l}{0.8559}          & 2.23            & \multicolumn{1}{l}{23.54}          & \multicolumn{1}{l}{0.8244}          & 4.40             & \multicolumn{1}{l}{22.75}          & \multicolumn{1}{l}{0.7916}          & 5.39            & \multicolumn{1}{l}{25.90}          & \multicolumn{1}{l}{0.8239}          & 4.01          \\ 
SRN \citep{tao2018scale}                       & \multicolumn{1}{l}{32.47}          & \multicolumn{1}{l}{0.8914}          & 2.08            & \multicolumn{1}{l}{24.13}          & \multicolumn{1}{l}{0.8656}          & 3.88            & \multicolumn{1}{l}{22.63}          & \multicolumn{1}{l}{0.7923}          & 5.45            & \multicolumn{1}{l}{26.41}                & \multicolumn{1}{l}{0.8498}          & 3.80         \\ 
 DeblurGANv2 \citep{kupyn2019deblurgan}                 & \multicolumn{1}{l}{30.26}          & \multicolumn{1}{l}{0.8504}          & 2.58            & \multicolumn{1}{l}{19.53}          & \multicolumn{1}{l}{0.7374}          & 6.29            & \multicolumn{1}{l}{22.56}          & \multicolumn{1}{l}{0.7735}          & 6.00              & \multicolumn{1}{l}{24.12}          & \multicolumn{1}{l}{0.7871}               & 4.96          \\ 
\textbf{DarkDeblurNet}             & \multicolumn{1}{l}{\textbf{34.56}} & \multicolumn{1}{l}{\textbf{0.9146}} & \textbf{1.78}   & \multicolumn{1}{l}{\textbf{24.88}} & \multicolumn{1}{l}{\textbf{0.8675}} & \textbf{3.62}   & \multicolumn{1}{l}{\textbf{24.09}} & \multicolumn{1}{l}{\textbf{0.8076}} & \textbf{4.60}    & \multicolumn{1}{l}{\textbf{27.84}} & \multicolumn{1}{l}{\textbf{0.8632}} & \textbf{3.33} \\ \hline
\end{tabular}}
\label{EDTAB}
\end{table}

As Table \ref{EDTAB} shows, the proposed DarkDeblurNet illustrates a significant improvement over existing methods while removing blurs in challenging low-light conditions. Notably, the performance gain is consistent in all evaluation matrices. Also, the learning-based methods are consistent on diverse data samples compared to their non-learning counterparts.

\begin{figure}[!htb]
\centering
\includegraphics[width=6.5cm,keepaspectratio]{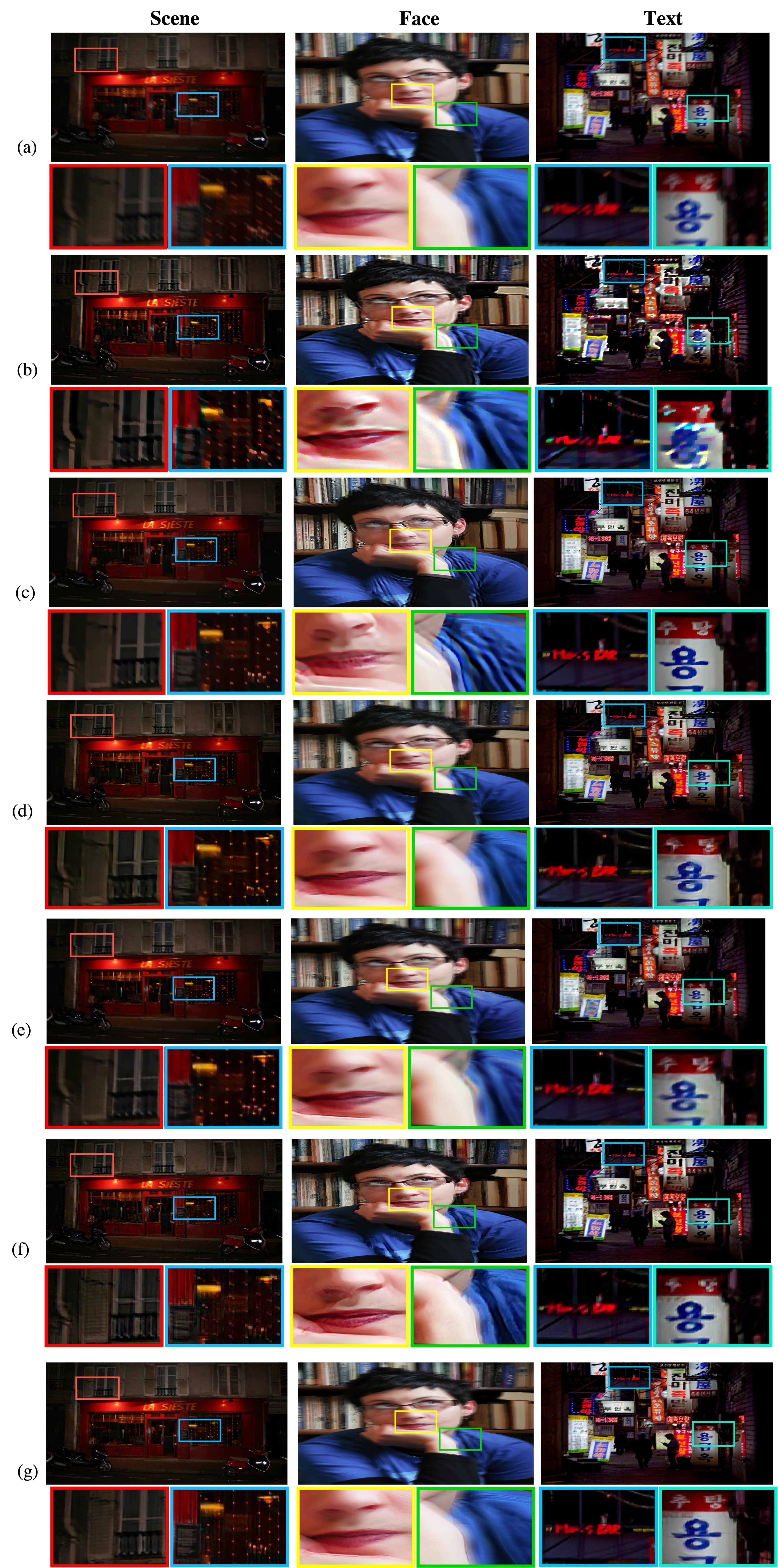}
\caption{ Qualitative comparison between the SOTA single-shot deblurring methods and proposed DarkDeblurNet on the ExDark dataset. It is visible that existing methods illustrate deficiencies while removing blurs from the image capture in low-light. (a) Blurry input image. (b) Result of LightStreaks . (c) Result of DarkChannel\citep{pan2016blind} . (d) Result of DeepDeblur \citep{nah2017deep}. (e) Result of SRN \citep{tao2018scale}. (f) Result of DeblurGANv2 \citep{kupyn2019deblurgan}. (g) DarkDeblurNet.}
\label{exDarkRes}
\end{figure}
 
 Figure \ref{exDarkRes} illustrates the qualitative comparison between the proposed DarkDebulNet and the SOTA methods in synthesized datasets. As the figure shows, the SOTA methods are prone to produce artifacts in low-light condition. Contrary, the proposed method demonstrates it's superiority over existing methods. It can produce visually pleasing images along with recovering far more details than its counterparts. In a nutshell, the proposed method can reduce artifacts and produces natural results.

\subsubsection{Low-light Deblurring (Real Data)}

Although all models were trained with synthesized data, it is salient to observe the performance of deep models in real-world data samples. Therefore, the performance of deep models has been studied on the proposed DarkShake dataset.

\begin{table}[!htb]
\centering
\caption{Quantitative comparison between SOTA methods and proposed DarkDeblurNet. A total of 100 image pairs from the DarkShake dataset were used to calculate the mean PSNR, SSIM, and DeltaE metrics. In all evaluation metrics, the DarkDeblurNet illustrated the consistency and outperformed the existing methods.}
\scalebox{.70}{\begin{tabular}{llll}
\toprule
\textbf{Method} & \textbf{PSNR $\uparrow$}  & \textbf{SSIM $\uparrow$}   & \textbf{DeltaE 	$\downarrow$} \\ \midrule
LightStreaks \citep{hu2014deblurring}       & 22.40           & 0.7299          & 5.18          \\
DarkChannel\citep{pan2016blind}  & 24.32          & 0.8214          & 3.58          \\
DeepDeblur \citep{nah2017deep}         & 24.54        & 0.7349          & 3.98        \\
SRN \citep{tao2018scale}         & 24.80        & 0.7410           & 3.92         \\
 DeblurGANv2 \citep{kupyn2019deblurgan}        & 23.32          & 0.7727          & 4.55          \\
\textbf{DarkDeblurNet} & \textbf{25.39} & \textbf{0.8401} & \textbf{3.75}      \\ \bottomrule
\end{tabular}}
\label{DSTab}

\end{table}

Table \ref{DSTab} illustrates the performance of deep models on the proposed DarkShake dataset. Here, the quantitative results are calculated on 25 blur-sharp image pairs. It is visible that the proposed model can outperform the existing methods in real-world data as well. It shows an improvement of 0.59 dB in PSNR metrics, 0.018 in SSIM metrics, and 0.17 in DeltaE metrics over SOTA methods.

 \begin{figure}[!htb]
\centering
\includegraphics[width=6.5cm,keepaspectratio]{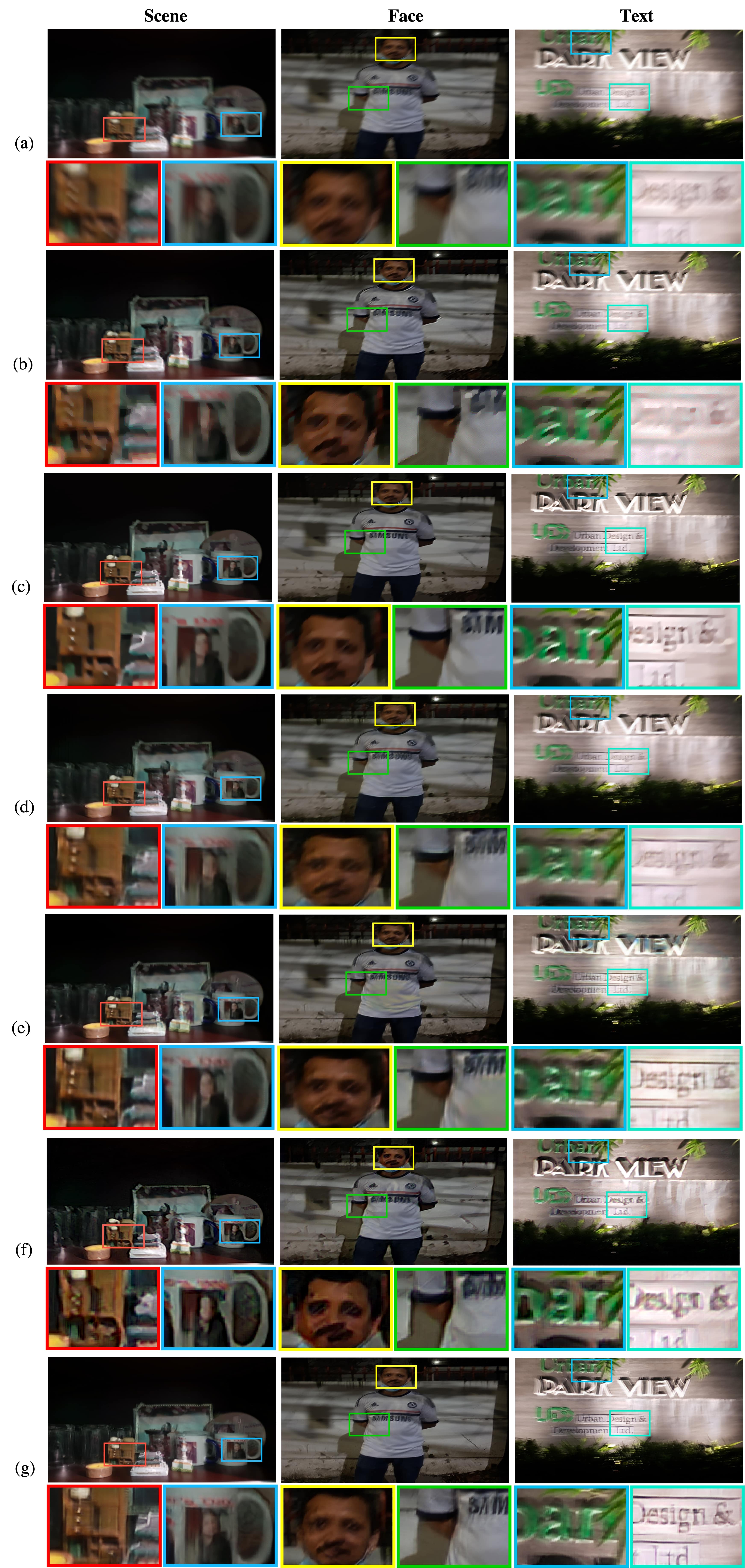}
\caption{Qualitative comparison between the SOTA methods and DarkDeblurNet on the proposed DarkShake dataset. The existing methods are prone to produce visually disturbing artifacts while removing blurs from real-world blurry images captured in low-light. (a) Blurry input image. (b) Result of LightStreaks . (c) Result of DarkChannel\citep{pan2016blind} . (d) Result of DeepDeblur \citep{nah2017deep}. (e) Result of SRN \citep{tao2018scale}. (f) Result of DeblurGANv2 \citep{kupyn2019deblurgan}. (g) DarkDeblurNet.}
\label{DarkShakeRes}
\end{figure}

 Figure \ref{DarkShakeRes} illustrates the visual comparison between the proposed DarkDebulNet and the SOTA methods in the proposed DarkShake dataset. On real-world data, the proposed DarkDeblurNet demonstrates consistency and depicts its superiority over SOTA methods. It is noteworthy that the proposed method can recover more details without producing any visually disturbing artifacts.

\subsubsection{Well-lit Deblurring }

The feasibility of the proposed method has been studied on well-lit conditions. Subsequently,  learning-based deblurring methods have been retrained with GoPro \citep{nah2017deep} and REDS \citep{nah2017deep} datasets. Later, the performance of all deblurring methods has evaluated with testing samples from the same datasets. Table. \ref{goredquant} illustrates the performance of SOTA deblurring methods on well-lit deblurring. It can be seen that the proposed method illustrates the consistency on the well-lit condition as well.

\begin{table}[!htb]
\centering

\caption{Quantitative comparison between SOTA methods GoPro \citep{nah2017deep} and REDS \citep{nah2017deep} datasets. The DarkDeblurNet can outperform the exiting methods for well-lit image deblurring.}
\scalebox{.45}{\begin{tabular}{llllllllll}
\hline
\multicolumn{1}{c}{\multirow{2}{*}{\textbf{Method}}} & \multicolumn{3}{l}{\textbf{GoPro Dataset}}                                                    & \multicolumn{3}{l}{\textbf{RED Dataset}}                                                      & \multicolumn{3}{l}{\textbf{Average}}                                                            \\ \cline{2-10} 
\multicolumn{1}{c}{}                                 & \multicolumn{1}{l}{\textbf{PSNR}}    & \multicolumn{1}{l}{\textbf{SSIM}}   & \textbf{DeltaE} & \multicolumn{1}{l}{\textbf{PSNR}}    & \multicolumn{1}{l}{\textbf{SSIM}}   & \textbf{DeltaE} & \multicolumn{1}{l}{\textbf{PSNR}}     & \multicolumn{1}{l}{\textbf{SSIM}}   & \textbf{DeltaE}  \\ \hline
LightStreaks \citep{hu2014deblurring}                                           & \multicolumn{1}{l}{21.25}            & \multicolumn{1}{l}{0.7378}          & 6.26            & \multicolumn{1}{l}{22.11}            & \multicolumn{1}{l}{0.7363}          & 5.58            & \multicolumn{1}{l}{21.68}             & \multicolumn{1}{l}{0.7371}         & 5.92             \\ 
DarkChannel\citep{pan2016blind}                                     & \multicolumn{1}{l}{24.75}            & \multicolumn{1}{l}{0.8730}           & 3.22            & \multicolumn{1}{l}{23.08}            & \multicolumn{1}{l}{0.8059}          & 3.99            & \multicolumn{1}{l}{23.92}            & \multicolumn{1}{l}{0.8395}         & 3.61            \\ 
DeepDeblur                                            & \multicolumn{1}{l}{26.03}          & \multicolumn{1}{l}{0.7936}          & 3.29           & \multicolumn{1}{l}{27.17}          & \multicolumn{1}{l}{0.7958}          & 3.18           & \multicolumn{1}{l}{26.60}          & \multicolumn{1}{l}{0.7947}          & 3.24           \\ 
SRN \citep{tao2018scale}                                             & \multicolumn{1}{l}{26.38}          & \multicolumn{1}{l}{0.8119}          & 3.13          & \multicolumn{1}{l}{27.77}          & \multicolumn{1}{l}{0.8148}          & 3.04           & \multicolumn{1}{l}{27.08}          & \multicolumn{1}{l}{0.8134}         & 3.08           \\ 

 DeblurGANv2 \citep{kupyn2019deblurgan}                                            & \multicolumn{1}{l}{24.33}          & \multicolumn{1}{l}{0.8208}          & 4.39          & \multicolumn{1}{l}{25.30}          & \multicolumn{1}{l}{0.8164}          & 4.47           & \multicolumn{1}{l}{24.82}          & \multicolumn{1}{l}{0.8186}         & 4.43           \\ 

\textbf{DarkDeblurNet}                                   & \multicolumn{1}{l}{\textbf{27.39}} & \multicolumn{1}{l}{0.8281}          & \textbf{2.75} & \multicolumn{1}{l}{\textbf{28.72}} & \multicolumn{1}{l}{0.8308}          & \textbf{2.66} & \multicolumn{1}{l}{\textbf{28.06}} & \multicolumn{1}{l}{0.8294}         & \textbf{2.71} \\ \hline
\end{tabular}}
\label{goredquant}
\end{table}

\begin{figure}[!htb]
\centering
\includegraphics[width=7.5cm,keepaspectratio]{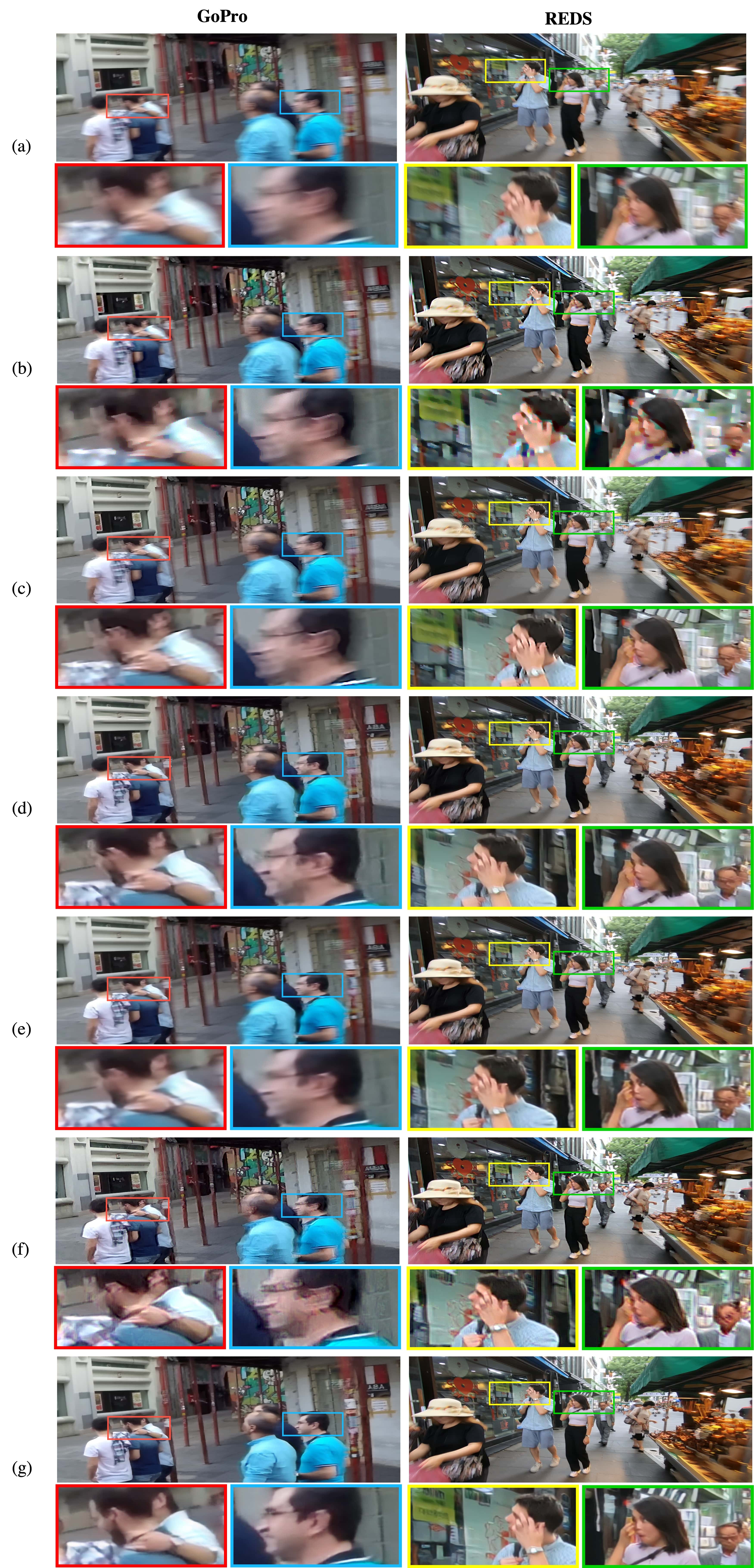}
\caption{Qualitative comparison between the SOTA methods and DarkDeblurNet on the GoPro \citep{nah2017deep} and REDS \citep{nah2017deep}. (a) Blurry input image. (b) Result of LightStreaks . (c) Result of DarkChannel\citep{pan2016blind} . (d) Result of DeepDeblur \citep{nah2017deep}. (e) Result of SRN \citep{tao2018scale}. (f) Result of DeblurGANv2 \citep{kupyn2019deblurgan}. (g) DarkDeblurNet.}
\label{GoREDSCom}
\end{figure}

In addition to the quantitative evaluation, the performance of the proposed method has been confirmed with a qualitative study. Figure \ref{GoREDSCom} depicts the visual comparison between existing deblurring methods. It is noticeable that the proposed method can produce sharper images without producing any visually disturbing artifacts. 

\subsection{Applications}

Digital cameras can produce blurry images due to numerous factors such as long-exposure settings, faster-moving objects, handshakes while holding the camera, etc., explicitly, with a single-shot setup. Regrettably, any extent of motion blurs drive camera hardware to capture the target scene in an unusable form. Apart from being a significant application of computer vision, the single-shot image deblurring can accelerate the performance of numerous real-world applications such as segmentation \citep{he2017mask,khan2020face}, detection \citep{redmon2018yolov3}, recognition \citep{thakare2018document}, microscopy \citep{han2017refocusing}, facial expression analysis \citep{ali2020artificial, naqvi2020deep}, 3D image analysis \citep{hanif2020novel}, medical imaging \citep{rundo2019medga}, space observation \citep{xu2017image}, etc.

Considering such widespread real-world applications, top computer vision societies like the computer vision foundation (CVF) arrange numerous completion and challenges in their top-tier conferences (i.e., CVPR, ECCV, ICCV, etc.) to encourage the development of deblurring solutions. However, most existing tracks and works for developing deblurring solutions are dedicated to normal lighting conditions. Contrarily, camera systems are commonly affected by motion blurs in low-light conditions due to the utilization of slower shutter speeds. This study addresses such inevitable motion blurs in challenging low-light conditions. 

Among the countless applications, this study illustrates the implication of single-shot low-light image deblurring in the three most widely used scenarios throughout the experiments.

\begin{figure}[!htb]
\centering
\includegraphics[width=\textwidth,keepaspectratio]{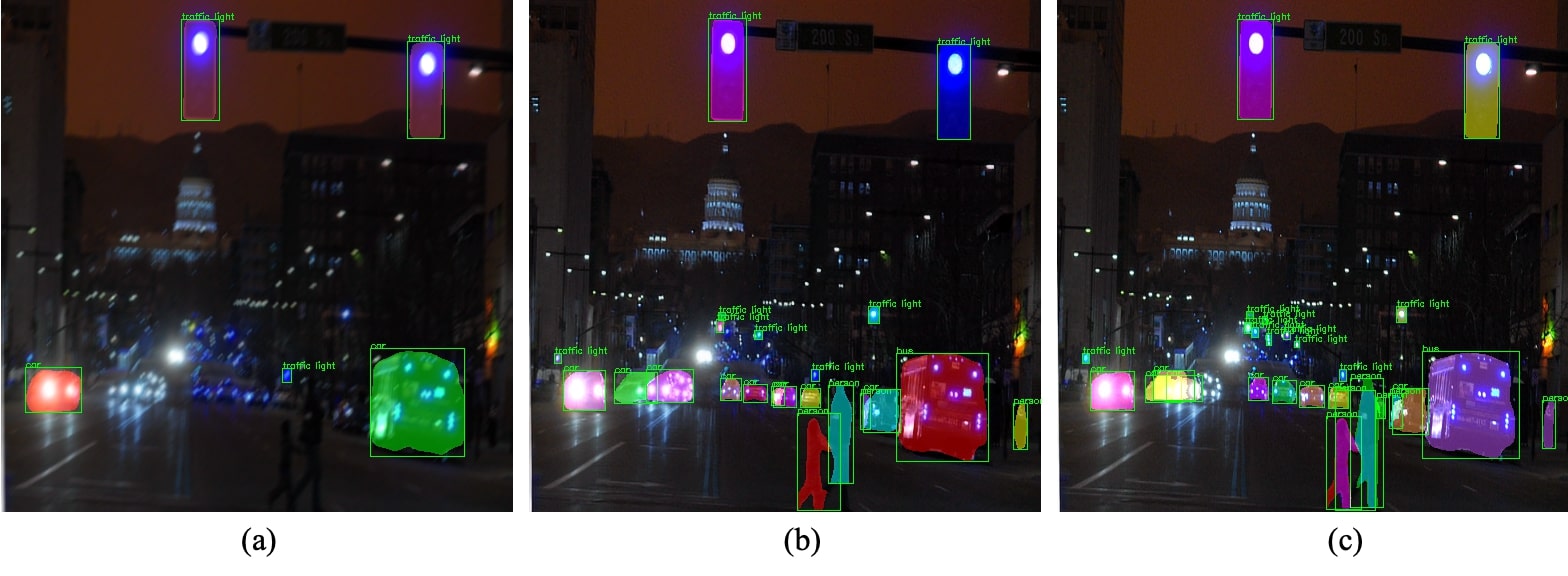}
\caption{ Performance improvement achieved by applying DarkDeblurNet with a segmentation method. (a) Blurry Image + Mask R-CNN \citep{he2017mask}. (b) DarkDeblurlNet + Mask R-CNN \citep{he2017mask}. (d) Reference Image + Mask R-CNN \citep{he2017mask}}
\label{segmentation}
\end{figure}

Object-Segmentation: Figure \ref{segmentation} illustrates the glims of performance improvement achieved in segmentation the utilization of proposed DarkDeblurNet. Here, the DarkDeblurNet is applied to a blurry image and then inference with a SOTA segmentation method known as Mask R-CNN. It is evident that the proposed method dramatically improves the performance of the segmentation method by allowing it to segment more objects than it performs with the blurry image.

\begin{figure}[ht]
\centering
\includegraphics[width=\textwidth,keepaspectratio]{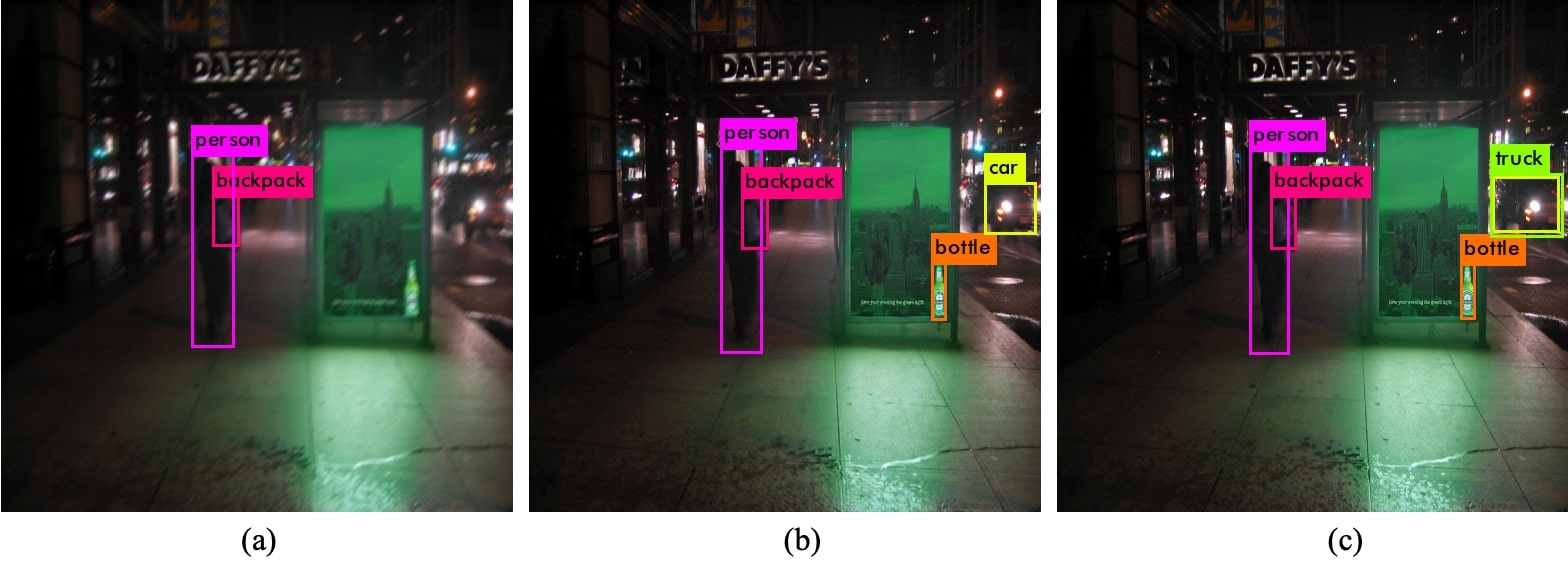}
\caption{ Performance improvement achieved by applying DarkDeblurNet with an object-detection method. (a) Blurry Image + YOLOv3 \citep{redmon2018yolov3}. (b) DarkDeblurlNet + YOLOv3 \citep{redmon2018yolov3}. (d) Reference Image + YOLOv3 \citep{redmon2018yolov3}.}
\label{detection}
\end{figure}

Object-Detection: Figure \ref{detection} demonstrates the performance improvement achieved in object segmentation by applying the proposed DarkDeblurNet. Here, to perform object detection, a SOTA method known as YOLOv3 \citep{redmon2018yolov3} is used. It is clear that the performance of the detection method is improved with a sharper image, which is restored through the proposed method.

\begin{figure}[!htb]
\centering
\includegraphics[width=\textwidth,keepaspectratio]{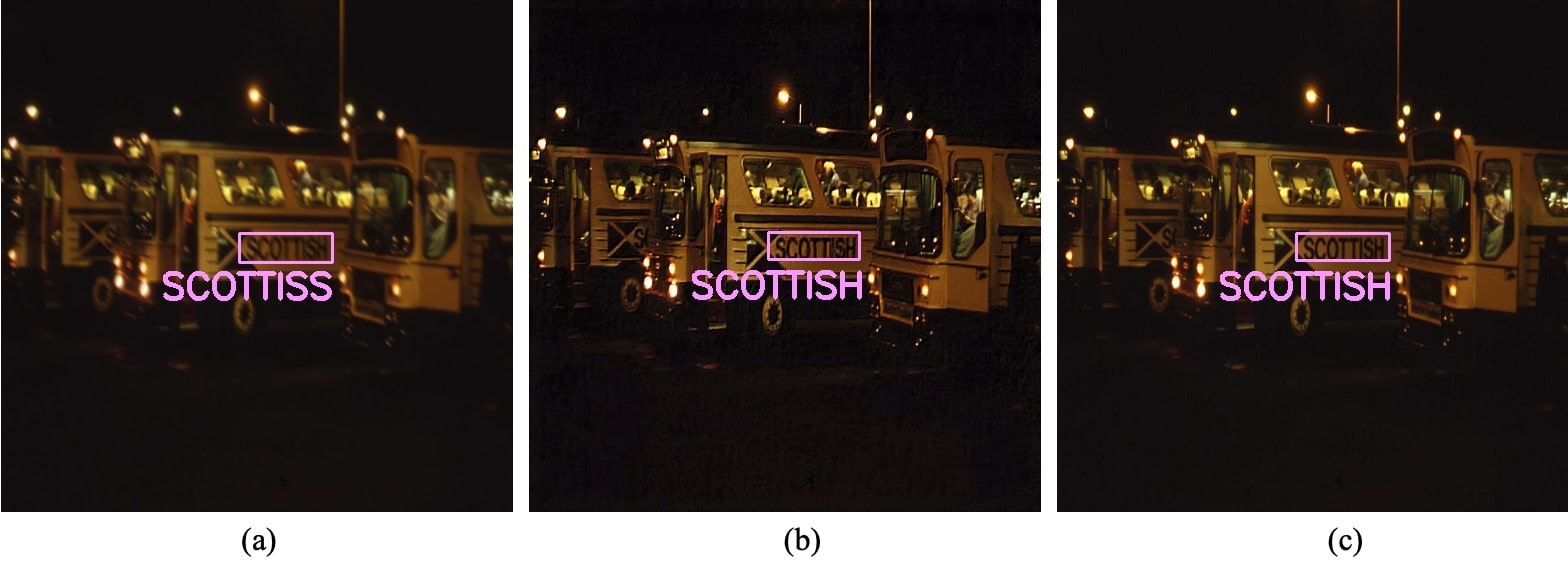}
\caption{ Performance improvement achieved by applying DarkDeblurNet with a text-recognition method. (a) Blurry Image + Tesseract \citep{thakare2018document}. (b) DarkDeblurlNet + Tesseract \citep{thakare2018document}. (d) Reference Image + Tesseract \citep{thakare2018document}.}
\label{recognition}
\end{figure}

Text-Recognition: Figure \ref{recognition} illustrates the performance improvement of an existing text recognition method by applying DarkDeblurNet on a blurry image. Here, the tesseract optical character recognition (OCR) framework \citep{thakare2018document} is applied for text recognition.

\subsection{Ablation Study}
The components proposed in this study have been verified with sophisticated experiments. As Table \ref{abTab} illustrates, the ablation experiment started with a baseline network architecture with a traditional dense block. Also, the baseline variant replaces the multi-term objective function with a simple L1-norm. Later, the proposed components like channel attention (CA) with short distance residual connection, contextual gate, and multi-term loss (ML) function are injected into the baseline architecture in a modular manner. Here, the baseline network with traditional dense block denoted in Table \ref{abTab} as DarkDeblurlNet$_{Base}$, DarkDeblurlNet$_{Base}$ with residual CA as DarkDeblurlNet$_{CA}$, and DarkDeblurlNet$_{CA}$ with CG as DarkDeblurlNet$_{CG}$. The network variant (proposed) with attention mechanisms and multi-term loss function has been denoted as DarkDeblurlNet.

\begin{table}[!htb]
\caption{Ablation study of the proposed method. The importance of the proposed component is verified with sophisticated experiments. In both synthesized and real-world blur removal,  each component plays a significant role in improving the performance of the DarkDeblurNet.}

\scalebox{.68}{\begin{tabular}{llllllllll}
\toprule
\multirow{2}{*}{\textbf{Model}} & \multirow{2}{*}{\textbf{CA}} & \multirow{2}{*}{\textbf{CG}} & \multirow{2}{*}{\textbf{ML}} & \multicolumn{3}{l}{\textbf{ExDark}}  & \multicolumn{3}{l}{\textbf{DarkShake}}    \\\cmidrule(){5-10}
                                &                              &                              &                               & \textbf{PSNR $\uparrow$}  & \textbf{SSIM $\uparrow$}   & \textbf{DeltaE $\downarrow$} & \textbf{PSNR $\uparrow$}  & \textbf{SSIM $\uparrow$}   & \textbf{DeltaE 	$\downarrow$} \\
\midrule
DarkDeblurlNet$_{Base}$              & \xmark                            & \xmark                             & \xmark                              & 25.37           & 0.7730         & 3.97            & 22.29              & 0.6711        & 5.86            \\
DarkDeblurlNet$_{CA}$               & \cmark                            & \xmark                             & \xmark                              & 28.29           & 0.7905        & 3.30             & 22.47              & 0.6238        & 6.26            \\
DarkDeblurlNet$_{CG}$              & \cmark                            & \cmark                            & \xmark                              & 31.56           & 0.8667        & 2.07            & 24.24              & 0.6904        & 3.96            \\
\textbf{DarkDeblurlNet}              & \cmark                            & \cmark                            & \cmark                             & \textbf{34.56} & \textbf{0.9146} & \textbf{1.78}   & \textbf{25.39} & \textbf{0.8401} & \textbf{3.75}   \\   
\bottomrule
\end{tabular}}
\label{abTab}

\end{table}

As Table \ref{abTab} shows, each proposed component has a significant role in a performance gain. Most notably,
the performance gap between the proposed DarkDeblurNet and its baseline variant (DarkDeblurlNet$_{Base}$) is immense. The proposed component such as short distance residual connection with CA, CG, and multi-term loss illustrates a substantial performance accretion of 9.19 dB on PSNR metrics, 0.1416 on SSIM metrics, and 2.19 on deltaE metrics for synthesized data. Also, the performance gain remains consistent on real-world data, where performance gain is calculated 2.51 dB on PSNR metrics, 0.0643 on SSIM metrics, and 1.96 deltaE metrics.

\begin{figure}[!htb]
\centering
\includegraphics[width=\textwidth,keepaspectratio]{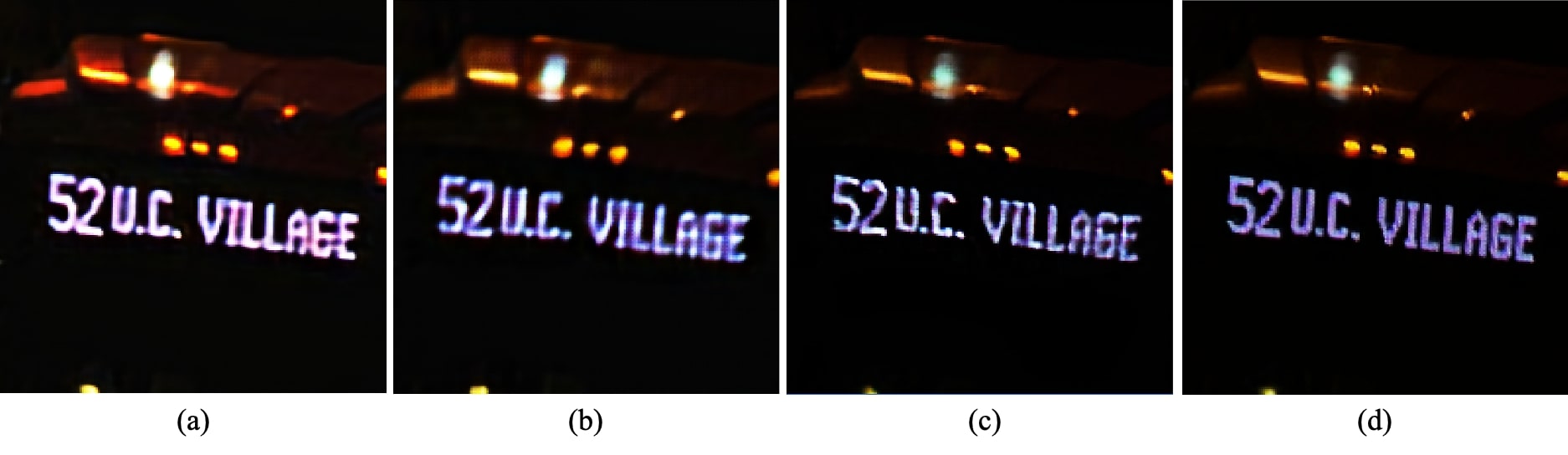}
\caption{ Qualitative verification of the proposed components on synthesized data. (a) DarkDeblurlNet$_{Base}$. (b) DarkDeblurlNet$_{DA}$. (d) DarkDeblurlNet$_{CG}$. (d) \textbf{DarkDeblurlNet (proposed)}. }
\label{ExDarkAB}
\end{figure}

\begin{figure}[!htb]
\centering
\includegraphics[width=\textwidth,keepaspectratio]{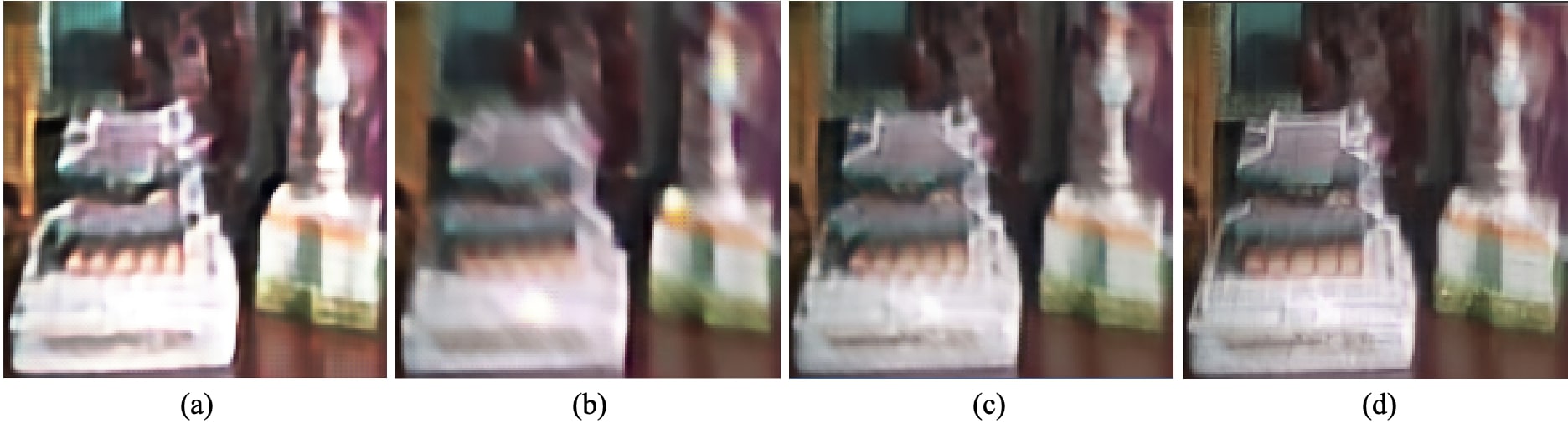}
\caption{ Qualitative verification of the proposed components on real-world blurry data. (a) DarkDeblurlNet$_{Base}$. (b) DarkDeblurlNet$_{DA}$. (d) DarkDeblurlNet$_{CG}$. (d) \textbf{DarkDeblurlNet (proposed)}. }
\label{DarkShakeAB}
\end{figure}

Figure \ref{ExDarkAB} and Figure \ref{DarkShakeAB} illustrate qualitative improvement achieved through the proposed components. It is noticeable that each of these components can enhance the perceptual quality in both synthesized and real-world data. Here, the  residual connection with CA in the dense-attention block can improve the global image quality. Similarly, CG enhances spatial information, while multi-term losses play a part in improving overall perceptual quality. In a nutshell, qualitative and quantitative comparison justifies the contribution of the proposed components for single-shot image deblurring in low-light conditions. 

\section{Discussion}
\label{discussion}
This section discusses the key findings, limitations, and future scope of the proposed work.

Network Architecture: The proposed DarkDeblurNet is presented as an FPN with a dense-attention block and contextual gates to learn feature interdependency precisely. Despite being deeper and wider, the complexity of the proposed network is calculated as 16.0 G floating-point operations (FLOPs). It is worth noting, the proposed network takes only 0.02s and 0.04s to infer on a single-shot image with a spatial dimension of $256 \times 256 \times 3$ and $512 \times 512 \times 3$, which is three times faster than the existing learning-based methods (i.e., DeepDeblur \citep{nah2017deep}, SRN \citep{tao2018scale}). Also, The proposed network is fully convolutional. Hence, it can take arbitrary image dimensions for inference.

\begin{figure}[!htb]
\centering
\includegraphics[width=\textwidth,keepaspectratio]{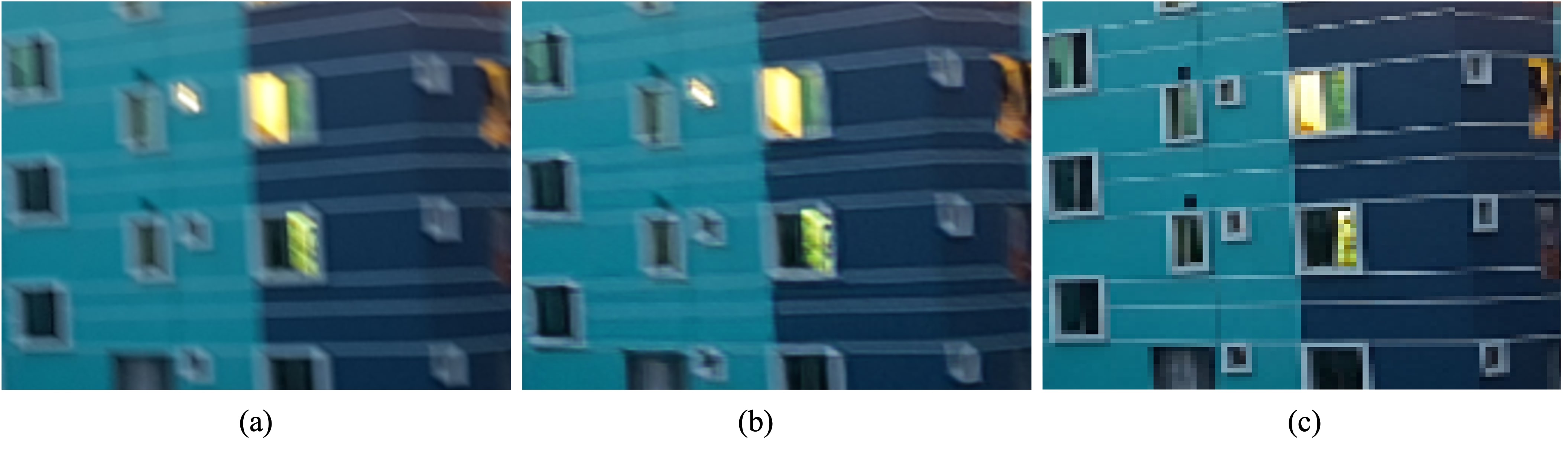}
\caption{ In extreme real-world cases, the proposed DarkDeblurNet can illustrate deficiencies. It has been suspected that lacking mostly arises due to the lacking real-world training data. (a) Input blurry image. (b) The output of DarkDeblurNet. (c) Reference (sharp) image.}
\label{flaws}
\end{figure}

Limitations: One of the limitations of the proposed study is identified as a lack of real-world low-light training data. Notably, the synthesized low-light blurry images differ from real-world blurry images and deteriorate the performance in real-world image samples (please see section \ref{results}). In some extreme cases, it can also compel the proposed network to demonstrate unwanted structural distortion, as shown in Figure. \ref{flaws}. 

Due to hardware limitations, the proposed study resized the large-dimensioned DarkShake image samples into the smaller spatial dimension to fit in the GPU memory. Thus, the compression artifacts can be observed in visualizing real-world examples.

Future Scope: The limitations of the proposed study lead to an interesting future direction. The findings of this study indicate that learning from real-world blur-sharp image pairs can be helpful for image deblurring in low-light conditions. Therefore, the proposed DarkShake dataset can be extended in such a way that it can use for training purposes. In another way, low-light image deblurring can be performed in an unsupervised manner. It can also resolve the training data limitations apart. 

It would be a very challenging but exciting task to combine image deblurring with other low-light image enhancement tasks (i.e., low-light to well-lit mapping). The performance of the proposed DarkDeblurNet can be studied to find the feasibility of joint image deblurring and low-light to well-lit mapping in the foreseeable future. Although the proposed DarkDeblurNet focuses on single-shot image deblurring, however,  the proposed work can be extended for multi-frame inference like video deblurring in a follow-up study.

\section{Conclusion}
\label{conclusion}
This study presented a learning-based single-shot image deblurring method specialized in low-light environments. The proposed DarkDeblurNet facilitated a novel dense-attention block in the feature pyramid structure for global image correction. Also, a contextual gating mechanism was used to propagate spatially enhanced features between different feature levels. The proposed architecture was optimized with a multi-term objective function. The feasibility of the proposed study was verified with sophisticated experiments. The proposed method illustrated superiority over SOTA methods in both synthesized and real-world testing. Apart from that, this study introduced a real-world blur-sharp image dataset, which can be used for low-light deblurring evaluations. It has planned to extend the proposed DarkShake dataset. Thus, it can be used for training to improve the performance of deep models. 

\section*{Acknowledgments}
This work was supported by the Sejong University Faculty Research Fund.

\bibliographystyle{elsarticle-harv}\biboptions{authoryear}

\begin{thebibliography}{83}
\expandafter\ifx\csname natexlab\endcsname\relax\def\natexlab#1{#1}\fi
\providecommand{\url}[1]{\texttt{#1}}
\providecommand{\href}[2]{#2}
\providecommand{\path}[1]{#1}
\providecommand{\DOIprefix}{doi:}
\providecommand{\ArXivprefix}{arXiv:}
\providecommand{\URLprefix}{URL: }
\providecommand{\Pubmedprefix}{pmid:}
\providecommand{\doi}[1]{\href{http://dx.doi.org/#1}{\path{#1}}}
\providecommand{\Pubmed}[1]{\href{pmid:#1}{\path{#1}}}
\providecommand{\bibinfo}[2]{#2}
\ifx\xfnm\relax \def\xfnm[#1]{\unskip,\space#1}\fi
\bibitem[{A~Sharif et~al.(2021)A~Sharif, Naqvi and Biswas}]{a2021beyond}
\bibinfo{author}{A~Sharif, S.}, \bibinfo{author}{Naqvi, R.A.},
  \bibinfo{author}{Biswas, M.}, \bibinfo{year}{2021}.
\newblock \bibinfo{title}{Beyond joint demosaicking and denoising: An image
  processing pipeline for a pixel-bin image sensor}, in:
  \bibinfo{booktitle}{Proceedings of the IEEE/CVF Conference on Computer Vision
  and Pattern Recognition}, pp. \bibinfo{pages}{233--242}.
\bibitem[{Aitken et~al.(2017)Aitken, Ledig, Theis, Caballero, Wang and
  Shi}]{aitken2017checkerboard}
\bibinfo{author}{Aitken, A.}, \bibinfo{author}{Ledig, C.},
  \bibinfo{author}{Theis, L.}, \bibinfo{author}{Caballero, J.},
  \bibinfo{author}{Wang, Z.}, \bibinfo{author}{Shi, W.}, \bibinfo{year}{2017}.
\newblock \bibinfo{title}{Checkerboard artifact free sub-pixel convolution: A
  note on sub-pixel convolution, resize convolution and convolution resize}.
\newblock \bibinfo{journal}{arXiv preprint arXiv:1707.02937} .
\bibitem[{Ali et~al.(2020)Ali, Ali, Ali, Draz, Majeed, Yasin, Ali and
  Haider}]{ali2020artificial}
\bibinfo{author}{Ali, G.}, \bibinfo{author}{Ali, A.}, \bibinfo{author}{Ali,
  F.}, \bibinfo{author}{Draz, U.}, \bibinfo{author}{Majeed, F.},
  \bibinfo{author}{Yasin, S.}, \bibinfo{author}{Ali, T.},
  \bibinfo{author}{Haider, N.}, \bibinfo{year}{2020}.
\newblock \bibinfo{title}{Artificial neural network based ensemble approach for
  multicultural facial expressions analysis}.
\newblock \bibinfo{journal}{IEEE Access} \bibinfo{volume}{8},
  \bibinfo{pages}{134950--134963}.
\bibitem[{Chakrabarti(2016)}]{chakrabarti2016neural}
\bibinfo{author}{Chakrabarti, A.}, \bibinfo{year}{2016}.
\newblock \bibinfo{title}{A neural approach to blind motion deblurring}, in:
  \bibinfo{booktitle}{European conference on computer vision},
  \bibinfo{organization}{Springer}. pp. \bibinfo{pages}{221--235}.
\bibitem[{Chatterjee et~al.(2011)Chatterjee, Joshi, Kang and
  Matsushita}]{chatterjee2011noise}
\bibinfo{author}{Chatterjee, P.}, \bibinfo{author}{Joshi, N.},
  \bibinfo{author}{Kang, S.B.}, \bibinfo{author}{Matsushita, Y.},
  \bibinfo{year}{2011}.
\newblock \bibinfo{title}{Noise suppression in low-light images through joint
  denoising and demosaicing}, in: \bibinfo{booktitle}{CVPR 2011},
  \bibinfo{organization}{IEEE}. pp. \bibinfo{pages}{321--328}.
\bibitem[{Cho and Lee(2009)}]{cho2009fast}
\bibinfo{author}{Cho, S.}, \bibinfo{author}{Lee, S.}, \bibinfo{year}{2009}.
\newblock \bibinfo{title}{Fast motion deblurring}, in: \bibinfo{booktitle}{ACM
  SIGGRAPH Asia 2009 papers}. \bibinfo{publisher}{Springer}, pp.
  \bibinfo{pages}{1--8}.
\bibitem[{Cort{\'e}s and Serratosa(2015)}]{cortes2015interactive}
\bibinfo{author}{Cort{\'e}s, X.}, \bibinfo{author}{Serratosa, F.},
  \bibinfo{year}{2015}.
\newblock \bibinfo{title}{An interactive method for the image alignment problem
  based on partially supervised correspondence}.
\newblock \bibinfo{journal}{Expert systems with applications}
  \bibinfo{volume}{42}, \bibinfo{pages}{179--192}.
\bibitem[{Dai et~al.(2019)Dai, Cai, Zhang, Xia and Zhang}]{dai2019second}
\bibinfo{author}{Dai, T.}, \bibinfo{author}{Cai, J.}, \bibinfo{author}{Zhang,
  Y.}, \bibinfo{author}{Xia, S.T.}, \bibinfo{author}{Zhang, L.},
  \bibinfo{year}{2019}.
\newblock \bibinfo{title}{Second-order attention network for single image
  super-resolution}, in: \bibinfo{booktitle}{Proceedings of the IEEE conference
  on computer vision and pattern recognition}, pp.
  \bibinfo{pages}{11065--11074}.
\bibitem[{Fergus et~al.(2006)Fergus, Singh, Hertzmann, Roweis and
  Freeman}]{fergus2006removing}
\bibinfo{author}{Fergus, R.}, \bibinfo{author}{Singh, B.},
  \bibinfo{author}{Hertzmann, A.}, \bibinfo{author}{Roweis, S.T.},
  \bibinfo{author}{Freeman, W.T.}, \bibinfo{year}{2006}.
\newblock \bibinfo{title}{Removing camera shake from a single photograph}, in:
  \bibinfo{booktitle}{ACM SIGGRAPH 2006 Papers}. \bibinfo{publisher}{ACM}, pp.
  \bibinfo{pages}{787--794}.
\bibitem[{Gai and Bao(2019)}]{gai2019new}
\bibinfo{author}{Gai, S.}, \bibinfo{author}{Bao, Z.}, \bibinfo{year}{2019}.
\newblock \bibinfo{title}{New image denoising algorithm via improved deep
  convolutional neural network with perceptive loss}.
\newblock \bibinfo{journal}{Expert Systems with Applications}
  \bibinfo{volume}{138}, \bibinfo{pages}{112815}.
\bibitem[{Gharbi et~al.(2017)Gharbi, Chen, Barron, Hasinoff and
  Durand}]{gharbi2017deep}
\bibinfo{author}{Gharbi, M.}, \bibinfo{author}{Chen, J.},
  \bibinfo{author}{Barron, J.T.}, \bibinfo{author}{Hasinoff, S.W.},
  \bibinfo{author}{Durand, F.}, \bibinfo{year}{2017}.
\newblock \bibinfo{title}{Deep bilateral learning for real-time image
  enhancement}.
\newblock \bibinfo{journal}{ACM Transactions on Graphics (TOG)}
  \bibinfo{volume}{36}, \bibinfo{pages}{1--12}.
\bibitem[{G{\'o}mez-Polo et~al.(2016)G{\'o}mez-Polo, Mu{\~n}oz, Luengo,
  Vicente, Galindo and Casado}]{gomez2016comparison}
\bibinfo{author}{G{\'o}mez-Polo, C.}, \bibinfo{author}{Mu{\~n}oz, M.P.},
  \bibinfo{author}{Luengo, M.C.L.}, \bibinfo{author}{Vicente, P.},
  \bibinfo{author}{Galindo, P.}, \bibinfo{author}{Casado, A.M.M.},
  \bibinfo{year}{2016}.
\newblock \bibinfo{title}{Comparison of the cielab and ciede2000 color
  difference formulas}.
\newblock \bibinfo{journal}{The journal of prosthetic dentistry}
  \bibinfo{volume}{115}, \bibinfo{pages}{65--70}.
\bibitem[{Goodfellow et~al.(2014)Goodfellow, Pouget-Abadie, Mirza, Xu,
  Warde-Farley, Ozair, Courville and Bengio}]{goodfellow2014generative}
\bibinfo{author}{Goodfellow, I.}, \bibinfo{author}{Pouget-Abadie, J.},
  \bibinfo{author}{Mirza, M.}, \bibinfo{author}{Xu, B.},
  \bibinfo{author}{Warde-Farley, D.}, \bibinfo{author}{Ozair, S.},
  \bibinfo{author}{Courville, A.}, \bibinfo{author}{Bengio, Y.},
  \bibinfo{year}{2014}.
\newblock \bibinfo{title}{Generative adversarial nets}, in:
  \bibinfo{booktitle}{Advances in neural information processing systems}, pp.
  \bibinfo{pages}{2672--2680}.
\bibitem[{Gupta et~al.(2010)Gupta, Joshi, Zitnick, Cohen and
  Curless}]{gupta2010single}
\bibinfo{author}{Gupta, A.}, \bibinfo{author}{Joshi, N.},
  \bibinfo{author}{Zitnick, C.L.}, \bibinfo{author}{Cohen, M.},
  \bibinfo{author}{Curless, B.}, \bibinfo{year}{2010}.
\newblock \bibinfo{title}{Single image deblurring using motion density
  functions}, in: \bibinfo{booktitle}{European Conference on Computer Vision},
  \bibinfo{organization}{Springer}. pp. \bibinfo{pages}{171--184}.
\bibitem[{Han and Yin(2017)}]{han2017refocusing}
\bibinfo{author}{Han, L.}, \bibinfo{author}{Yin, Z.}, \bibinfo{year}{2017}.
\newblock \bibinfo{title}{Refocusing phase contrast microscopy images}, in:
  \bibinfo{booktitle}{International Conference on Medical Image Computing and
  Computer-Assisted Intervention}, \bibinfo{organization}{Springer}. pp.
  \bibinfo{pages}{65--74}.
\bibitem[{Hanif et~al.(2020)Hanif, Naqvi, Abbas, Khan and
  Iqbal}]{hanif2020novel}
\bibinfo{author}{Hanif, M.}, \bibinfo{author}{Naqvi, R.A.},
  \bibinfo{author}{Abbas, S.}, \bibinfo{author}{Khan, M.A.},
  \bibinfo{author}{Iqbal, N.}, \bibinfo{year}{2020}.
\newblock \bibinfo{title}{A novel and efficient 3d multiple images encryption
  scheme based on chaotic systems and swapping operations}.
\newblock \bibinfo{journal}{IEEE Access} \bibinfo{volume}{8},
  \bibinfo{pages}{123536--123555}.
\bibitem[{He et~al.(2017)He, Gkioxari, Doll{\'a}r and Girshick}]{he2017mask}
\bibinfo{author}{He, K.}, \bibinfo{author}{Gkioxari, G.},
  \bibinfo{author}{Doll{\'a}r, P.}, \bibinfo{author}{Girshick, R.},
  \bibinfo{year}{2017}.
\newblock \bibinfo{title}{Mask r-cnn}, in: \bibinfo{booktitle}{Proceedings of
  the IEEE international conference on computer vision}, pp.
  \bibinfo{pages}{2961--2969}.
\bibitem[{He et~al.(2016)He, Zhang, Ren and Sun}]{he2016deep}
\bibinfo{author}{He, K.}, \bibinfo{author}{Zhang, X.}, \bibinfo{author}{Ren,
  S.}, \bibinfo{author}{Sun, J.}, \bibinfo{year}{2016}.
\newblock \bibinfo{title}{Deep residual learning for image recognition}, in:
  \bibinfo{booktitle}{Proceedings of the IEEE conference on computer vision and
  pattern recognition}, pp. \bibinfo{pages}{770--778}.
\bibitem[{Hu et~al.(2018)Hu, Shen and Sun}]{hu2018squeeze}
\bibinfo{author}{Hu, J.}, \bibinfo{author}{Shen, L.}, \bibinfo{author}{Sun,
  G.}, \bibinfo{year}{2018}.
\newblock \bibinfo{title}{Squeeze-and-excitation networks}, in:
  \bibinfo{booktitle}{Proceedings of the IEEE conference on computer vision and
  pattern recognition}, pp. \bibinfo{pages}{7132--7141}.
\bibitem[{Hu et~al.(2014)Hu, Cho, Wang and Yang}]{hu2014deblurring}
\bibinfo{author}{Hu, Z.}, \bibinfo{author}{Cho, S.}, \bibinfo{author}{Wang,
  J.}, \bibinfo{author}{Yang, M.H.}, \bibinfo{year}{2014}.
\newblock \bibinfo{title}{Deblurring low-light images with light streaks}, in:
  \bibinfo{booktitle}{Proceedings of the IEEE Conference on Computer Vision and
  Pattern Recognition}, pp. \bibinfo{pages}{3382--3389}.
\bibitem[{Huang et~al.(2017)Huang, Liu, Van Der~Maaten and
  Weinberger}]{huang2017densely}
\bibinfo{author}{Huang, G.}, \bibinfo{author}{Liu, Z.}, \bibinfo{author}{Van
  Der~Maaten, L.}, \bibinfo{author}{Weinberger, K.Q.}, \bibinfo{year}{2017}.
\newblock \bibinfo{title}{Densely connected convolutional networks}, in:
  \bibinfo{booktitle}{Proceedings of the IEEE conference on computer vision and
  pattern recognition}, pp. \bibinfo{pages}{4700--4708}.
\bibitem[{Ignatov et~al.(2017)Ignatov, Kobyshev, Timofte, Vanhoey and
  Van~Gool}]{ignatov2017dslr}
\bibinfo{author}{Ignatov, A.}, \bibinfo{author}{Kobyshev, N.},
  \bibinfo{author}{Timofte, R.}, \bibinfo{author}{Vanhoey, K.},
  \bibinfo{author}{Van~Gool, L.}, \bibinfo{year}{2017}.
\newblock \bibinfo{title}{Dslr-quality photos on mobile devices with deep
  convolutional networks}, in: \bibinfo{booktitle}{Proceedings of the IEEE
  International Conference on Computer Vision}, pp.
  \bibinfo{pages}{3277--3285}.
\bibitem[{Jiang et~al.(2019)Jiang, Gong, Liu, Cheng, Fang, Shen, Yang, Zhou and
  Wang}]{jiang2019enlightengan}
\bibinfo{author}{Jiang, Y.}, \bibinfo{author}{Gong, X.}, \bibinfo{author}{Liu,
  D.}, \bibinfo{author}{Cheng, Y.}, \bibinfo{author}{Fang, C.},
  \bibinfo{author}{Shen, X.}, \bibinfo{author}{Yang, J.},
  \bibinfo{author}{Zhou, P.}, \bibinfo{author}{Wang, Z.}, \bibinfo{year}{2019}.
\newblock \bibinfo{title}{Enlightengan: Deep light enhancement without paired
  supervision}.
\newblock \bibinfo{journal}{arXiv preprint arXiv:1906.06972} .
\bibitem[{Jobson et~al.(1997)Jobson, Rahman and Woodell}]{jobson1997multiscale}
\bibinfo{author}{Jobson, D.J.}, \bibinfo{author}{Rahman, Z.u.},
  \bibinfo{author}{Woodell, G.A.}, \bibinfo{year}{1997}.
\newblock \bibinfo{title}{A multiscale retinex for bridging the gap between
  color images and the human observation of scenes}.
\newblock \bibinfo{journal}{IEEE Transactions on Image processing}
  \bibinfo{volume}{6}, \bibinfo{pages}{965--976}.
\bibitem[{Johnson et~al.(2016)Johnson, Alahi and
  Fei-Fei}]{johnson2016perceptual}
\bibinfo{author}{Johnson, J.}, \bibinfo{author}{Alahi, A.},
  \bibinfo{author}{Fei-Fei, L.}, \bibinfo{year}{2016}.
\newblock \bibinfo{title}{Perceptual losses for real-time style transfer and
  super-resolution}, in: \bibinfo{booktitle}{European conference on computer
  vision}, \bibinfo{organization}{Springer}. pp. \bibinfo{pages}{694--711}.
\bibitem[{Khan et~al.(2020)Khan, Khan, Ahmad, Ali and Kwak}]{khan2020face}
\bibinfo{author}{Khan, K.}, \bibinfo{author}{Khan, R.U.},
  \bibinfo{author}{Ahmad, K.}, \bibinfo{author}{Ali, F.},
  \bibinfo{author}{Kwak, K.S.}, \bibinfo{year}{2020}.
\newblock \bibinfo{title}{Face segmentation: A journey from classical to deep
  learning paradigm, approaches, trends, and directions}.
\newblock \bibinfo{journal}{IEEE Access} \bibinfo{volume}{8},
  \bibinfo{pages}{58683--58699}.
\bibitem[{Kingma and Ba(2014)}]{kingma2014adam}
\bibinfo{author}{Kingma, D.P.}, \bibinfo{author}{Ba, J.}, \bibinfo{year}{2014}.
\newblock \bibinfo{title}{Adam: A method for stochastic optimization}.
\newblock \bibinfo{journal}{arXiv preprint arXiv:1412.6980} .
\bibitem[{Kirillov et~al.(2019)Kirillov, Girshick, He and
  Doll{\'a}r}]{kirillov2019panoptic}
\bibinfo{author}{Kirillov, A.}, \bibinfo{author}{Girshick, R.},
  \bibinfo{author}{He, K.}, \bibinfo{author}{Doll{\'a}r, P.},
  \bibinfo{year}{2019}.
\newblock \bibinfo{title}{Panoptic feature pyramid networks}, in:
  \bibinfo{booktitle}{Proceedings of the IEEE Conference on Computer Vision and
  Pattern Recognition}, pp. \bibinfo{pages}{6399--6408}.
\bibitem[{Krishnan et~al.(2011)Krishnan, Tay and Fergus}]{krishnan2011blind}
\bibinfo{author}{Krishnan, D.}, \bibinfo{author}{Tay, T.},
  \bibinfo{author}{Fergus, R.}, \bibinfo{year}{2011}.
\newblock \bibinfo{title}{Blind deconvolution using a normalized sparsity
  measure}, in: \bibinfo{booktitle}{CVPR 2011}, \bibinfo{organization}{IEEE}.
  pp. \bibinfo{pages}{233--240}.
\bibitem[{Kupyn et~al.(2018)Kupyn, Budzan, Mykhailych, Mishkin and
  Matas}]{kupyn2018deblurgan}
\bibinfo{author}{Kupyn, O.}, \bibinfo{author}{Budzan, V.},
  \bibinfo{author}{Mykhailych, M.}, \bibinfo{author}{Mishkin, D.},
  \bibinfo{author}{Matas, J.}, \bibinfo{year}{2018}.
\newblock \bibinfo{title}{Deblurgan: Blind motion deblurring using conditional
  adversarial networks}, in: \bibinfo{booktitle}{Proceedings of the IEEE
  conference on computer vision and pattern recognition}, pp.
  \bibinfo{pages}{8183--8192}.
\bibitem[{Kupyn et~al.(2019)Kupyn, Martyniuk, Wu and Wang}]{kupyn2019deblurgan}
\bibinfo{author}{Kupyn, O.}, \bibinfo{author}{Martyniuk, T.},
  \bibinfo{author}{Wu, J.}, \bibinfo{author}{Wang, Z.}, \bibinfo{year}{2019}.
\newblock \bibinfo{title}{Deblurgan-v2: Deblurring (orders-of-magnitude) faster
  and better}, in: \bibinfo{booktitle}{Proceedings of the IEEE International
  Conference on Computer Vision}, pp. \bibinfo{pages}{8878--8887}.
\bibitem[{Lai et~al.(2017)Lai, Huang, Ahuja and Yang}]{lai2017deep}
\bibinfo{author}{Lai, W.S.}, \bibinfo{author}{Huang, J.B.},
  \bibinfo{author}{Ahuja, N.}, \bibinfo{author}{Yang, M.H.},
  \bibinfo{year}{2017}.
\newblock \bibinfo{title}{Deep laplacian pyramid networks for fast and accurate
  super-resolution}, in: \bibinfo{booktitle}{Proceedings of the IEEE conference
  on computer vision and pattern recognition}, pp. \bibinfo{pages}{624--632}.
\bibitem[{Lai et~al.(2016)Lai, Huang, Hu, Ahuja and Yang}]{lai2016comparative}
\bibinfo{author}{Lai, W.S.}, \bibinfo{author}{Huang, J.B.},
  \bibinfo{author}{Hu, Z.}, \bibinfo{author}{Ahuja, N.}, \bibinfo{author}{Yang,
  M.H.}, \bibinfo{year}{2016}.
\newblock \bibinfo{title}{A comparative study for single image blind
  deblurring}, in: \bibinfo{booktitle}{Proceedings of the IEEE Conference on
  Computer Vision and Pattern Recognition}, pp. \bibinfo{pages}{1701--1709}.
\bibitem[{Land(1977)}]{land1977retinex}
\bibinfo{author}{Land, E.H.}, \bibinfo{year}{1977}.
\newblock \bibinfo{title}{The retinex theory of color vision}.
\newblock \bibinfo{journal}{Scientific american} \bibinfo{volume}{237},
  \bibinfo{pages}{108--129}.
\bibitem[{Ledig et~al.(2017)Ledig, Theis, Husz{\'a}r, Caballero, Cunningham,
  Acosta, Aitken, Tejani, Totz, Wang et~al.}]{ledig2017photo}
\bibinfo{author}{Ledig, C.}, \bibinfo{author}{Theis, L.},
  \bibinfo{author}{Husz{\'a}r, F.}, \bibinfo{author}{Caballero, J.},
  \bibinfo{author}{Cunningham, A.}, \bibinfo{author}{Acosta, A.},
  \bibinfo{author}{Aitken, A.}, \bibinfo{author}{Tejani, A.},
  \bibinfo{author}{Totz, J.}, \bibinfo{author}{Wang, Z.}, et~al.,
  \bibinfo{year}{2017}.
\newblock \bibinfo{title}{Photo-realistic single image super-resolution using a
  generative adversarial network}, in: \bibinfo{booktitle}{Proceedings of the
  IEEE conference on computer vision and pattern recognition}, pp.
  \bibinfo{pages}{4681--4690}.
\bibitem[{Levin et~al.(2011a)Levin, Weiss, Durand and
  Freeman}]{levin2011efficient}
\bibinfo{author}{Levin, A.}, \bibinfo{author}{Weiss, Y.},
  \bibinfo{author}{Durand, F.}, \bibinfo{author}{Freeman, W.T.},
  \bibinfo{year}{2011}a.
\newblock \bibinfo{title}{Efficient marginal likelihood optimization in blind
  deconvolution}, in: \bibinfo{booktitle}{CVPR 2011},
  \bibinfo{organization}{IEEE}. pp. \bibinfo{pages}{2657--2664}.
\bibitem[{Levin et~al.(2011b)Levin, Weiss, Durand and
  Freeman}]{levin2011understanding}
\bibinfo{author}{Levin, A.}, \bibinfo{author}{Weiss, Y.},
  \bibinfo{author}{Durand, F.}, \bibinfo{author}{Freeman, W.T.},
  \bibinfo{year}{2011}b.
\newblock \bibinfo{title}{Understanding blind deconvolution algorithms}.
\newblock \bibinfo{journal}{IEEE transactions on pattern analysis and machine
  intelligence} \bibinfo{volume}{33}, \bibinfo{pages}{2354--2367}.
\bibitem[{Lin et~al.(2017)Lin, Doll{\'a}r, Girshick, He, Hariharan and
  Belongie}]{lin2017feature}
\bibinfo{author}{Lin, T.Y.}, \bibinfo{author}{Doll{\'a}r, P.},
  \bibinfo{author}{Girshick, R.}, \bibinfo{author}{He, K.},
  \bibinfo{author}{Hariharan, B.}, \bibinfo{author}{Belongie, S.},
  \bibinfo{year}{2017}.
\newblock \bibinfo{title}{Feature pyramid networks for object detection}, in:
  \bibinfo{booktitle}{Proceedings of the IEEE conference on computer vision and
  pattern recognition}, pp. \bibinfo{pages}{2117--2125}.
\bibitem[{Liu et~al.(2014)Liu, Wang and Wang}]{liu2014effect}
\bibinfo{author}{Liu, H.}, \bibinfo{author}{Wang, Y.}, \bibinfo{author}{Wang,
  L.}, \bibinfo{year}{2014}.
\newblock \bibinfo{title}{The effect of light conditions on
  photoplethysmographic image acquisition using a commercial camera}.
\newblock \bibinfo{journal}{IEEE journal of translational engineering in health
  and medicine} \bibinfo{volume}{2}, \bibinfo{pages}{1--11}.
\bibitem[{Liu and Han(2018)}]{liu2018deep}
\bibinfo{author}{Liu, N.}, \bibinfo{author}{Han, J.}, \bibinfo{year}{2018}.
\newblock \bibinfo{title}{A deep spatial contextual long-term recurrent
  convolutional network for saliency detection}.
\newblock \bibinfo{journal}{IEEE Transactions on Image Processing}
  \bibinfo{volume}{27}, \bibinfo{pages}{3264--3274}.
\bibitem[{Liu et~al.(2020)Liu, Xiao, Fan, Zhao, Tang and
  Liu}]{liu2020importance}
\bibinfo{author}{Liu, P.}, \bibinfo{author}{Xiao, T.}, \bibinfo{author}{Fan,
  C.}, \bibinfo{author}{Zhao, W.}, \bibinfo{author}{Tang, X.},
  \bibinfo{author}{Liu, H.}, \bibinfo{year}{2020}.
\newblock \bibinfo{title}{Importance-weighted conditional adversarial network
  for unsupervised domain adaptation}.
\newblock \bibinfo{journal}{Expert Systems with Applications} ,
  \bibinfo{pages}{113404}.
\bibitem[{Loh and Chan(2019)}]{Exdark}
\bibinfo{author}{Loh, Y.P.}, \bibinfo{author}{Chan, C.S.},
  \bibinfo{year}{2019}.
\newblock \bibinfo{title}{Getting to know low-light images with the exclusively
  dark dataset}.
\newblock \bibinfo{journal}{Computer Vision and Image Understanding}
  \bibinfo{volume}{178}, \bibinfo{pages}{30--42}.
\newblock \DOIprefix\doi{https://doi.org/10.1016/j.cviu.2018.10.010}.
\bibitem[{Lore et~al.(2017)Lore, Akintayo and Sarkar}]{lore2017llnet}
\bibinfo{author}{Lore, K.G.}, \bibinfo{author}{Akintayo, A.},
  \bibinfo{author}{Sarkar, S.}, \bibinfo{year}{2017}.
\newblock \bibinfo{title}{Llnet: A deep autoencoder approach to natural
  low-light image enhancement}.
\newblock \bibinfo{journal}{Pattern Recognition} \bibinfo{volume}{61},
  \bibinfo{pages}{650--662}.
\bibitem[{Lowe(2004)}]{lowe2004distinctive}
\bibinfo{author}{Lowe, D.G.}, \bibinfo{year}{2004}.
\newblock \bibinfo{title}{Distinctive image features from scale-invariant
  keypoints}.
\newblock \bibinfo{journal}{International journal of computer vision}
  \bibinfo{volume}{60}, \bibinfo{pages}{91--110}.
\bibitem[{Lv et~al.(2019)Lv, Li and Lu}]{lv2019attention}
\bibinfo{author}{Lv, F.}, \bibinfo{author}{Li, Y.}, \bibinfo{author}{Lu, F.},
  \bibinfo{year}{2019}.
\newblock \bibinfo{title}{Attention guided low-light image enhancement with a
  large scale low-light simulation dataset}.
\newblock \bibinfo{journal}{arXiv preprint arXiv:1908.00682} .
\bibitem[{Mirza and Osindero(2014)}]{mirza2014conditional}
\bibinfo{author}{Mirza, M.}, \bibinfo{author}{Osindero, S.},
  \bibinfo{year}{2014}.
\newblock \bibinfo{title}{Conditional generative adversarial nets}.
\newblock \bibinfo{journal}{arXiv preprint arXiv:1411.1784} .
\bibitem[{Money and Kang(2008)}]{money2008total}
\bibinfo{author}{Money, J.H.}, \bibinfo{author}{Kang, S.H.},
  \bibinfo{year}{2008}.
\newblock \bibinfo{title}{Total variation minimizing blind deconvolution with
  shock filter reference}.
\newblock \bibinfo{journal}{Image and Vision Computing} \bibinfo{volume}{26},
  \bibinfo{pages}{302--314}.
\bibitem[{Nah et~al.(2017)Nah, Hyun~Kim and Mu~Lee}]{nah2017deep}
\bibinfo{author}{Nah, S.}, \bibinfo{author}{Hyun~Kim, T.},
  \bibinfo{author}{Mu~Lee, K.}, \bibinfo{year}{2017}.
\newblock \bibinfo{title}{Deep multi-scale convolutional neural network for
  dynamic scene deblurring}, in: \bibinfo{booktitle}{Proceedings of the IEEE
  Conference on Computer Vision and Pattern Recognition}, pp.
  \bibinfo{pages}{3883--3891}.
\bibitem[{Naqvi et~al.(2020)Naqvi, Arsalan, Rehman, Rehman, Loh and
  Paul}]{naqvi2020deep}
\bibinfo{author}{Naqvi, R.A.}, \bibinfo{author}{Arsalan, M.},
  \bibinfo{author}{Rehman, A.}, \bibinfo{author}{Rehman, A.U.},
  \bibinfo{author}{Loh, W.K.}, \bibinfo{author}{Paul, A.},
  \bibinfo{year}{2020}.
\newblock \bibinfo{title}{Deep learning-based drivers emotion classification
  system in time series data for remote applications}.
\newblock \bibinfo{journal}{Remote Sensing} \bibinfo{volume}{12},
  \bibinfo{pages}{587}.
\bibitem[{Pan et~al.(2016)Pan, Sun, Pfister and Yang}]{pan2016blind}
\bibinfo{author}{Pan, J.}, \bibinfo{author}{Sun, D.}, \bibinfo{author}{Pfister,
  H.}, \bibinfo{author}{Yang, M.H.}, \bibinfo{year}{2016}.
\newblock \bibinfo{title}{Blind image deblurring using dark channel prior}, in:
  \bibinfo{booktitle}{Proceedings of the IEEE Conference on Computer Vision and
  Pattern Recognition}, pp. \bibinfo{pages}{1628--1636}.
\bibitem[{Perrone and Favaro(2015)}]{perrone2015clearer}
\bibinfo{author}{Perrone, D.}, \bibinfo{author}{Favaro, P.},
  \bibinfo{year}{2015}.
\newblock \bibinfo{title}{A clearer picture of total variation blind
  deconvolution}.
\newblock \bibinfo{journal}{IEEE transactions on pattern analysis and machine
  intelligence} \bibinfo{volume}{38}, \bibinfo{pages}{1041--1055}.
\bibitem[{Perrone and Favaro(2016)}]{perrone2016logarithmic}
\bibinfo{author}{Perrone, D.}, \bibinfo{author}{Favaro, P.},
  \bibinfo{year}{2016}.
\newblock \bibinfo{title}{A logarithmic image prior for blind deconvolution}.
\newblock \bibinfo{journal}{International journal of computer vision}
  \bibinfo{volume}{117}, \bibinfo{pages}{159--172}.
\bibitem[{Pizer et~al.(1987)Pizer, Amburn, Austin, Cromartie, Geselowitz,
  Greer, ter Haar~Romeny, Zimmerman and Zuiderveld}]{pizer1987adaptive}
\bibinfo{author}{Pizer, S.M.}, \bibinfo{author}{Amburn, E.P.},
  \bibinfo{author}{Austin, J.D.}, \bibinfo{author}{Cromartie, R.},
  \bibinfo{author}{Geselowitz, A.}, \bibinfo{author}{Greer, T.},
  \bibinfo{author}{ter Haar~Romeny, B.}, \bibinfo{author}{Zimmerman, J.B.},
  \bibinfo{author}{Zuiderveld, K.}, \bibinfo{year}{1987}.
\newblock \bibinfo{title}{Adaptive histogram equalization and its variations}.
\newblock \bibinfo{journal}{Computer vision, graphics, and image processing}
  \bibinfo{volume}{39}, \bibinfo{pages}{355--368}.
\bibitem[{Pytorch(2016)}]{pytorch}
\bibinfo{author}{Pytorch}, \bibinfo{year}{2016}.
\newblock \bibinfo{title}{{PyTorch Framework} code}.
\newblock \bibinfo{howpublished}{\url{https://pytorch.org/}}.
\newblock \bibinfo{note}{Accessed: 2020-08-24}.
\bibitem[{Redmon and Farhadi(2018)}]{redmon2018yolov3}
\bibinfo{author}{Redmon, J.}, \bibinfo{author}{Farhadi, A.},
  \bibinfo{year}{2018}.
\newblock \bibinfo{title}{Yolov3: An incremental improvement}.
\newblock \bibinfo{journal}{arXiv preprint arXiv:1804.02767} .
\bibitem[{Ronneberger et~al.(2015)Ronneberger, Fischer and
  Brox}]{ronneberger2015u}
\bibinfo{author}{Ronneberger, O.}, \bibinfo{author}{Fischer, P.},
  \bibinfo{author}{Brox, T.}, \bibinfo{year}{2015}.
\newblock \bibinfo{title}{U-net: Convolutional networks for biomedical image
  segmentation}, in: \bibinfo{booktitle}{International Conference on Medical
  image computing and computer-assisted intervention},
  \bibinfo{organization}{Springer}. pp. \bibinfo{pages}{234--241}.
\bibitem[{Rundo et~al.(2019)Rundo, Tangherloni, Nobile, Militello, Besozzi,
  Mauri and Cazzaniga}]{rundo2019medga}
\bibinfo{author}{Rundo, L.}, \bibinfo{author}{Tangherloni, A.},
  \bibinfo{author}{Nobile, M.S.}, \bibinfo{author}{Militello, C.},
  \bibinfo{author}{Besozzi, D.}, \bibinfo{author}{Mauri, G.},
  \bibinfo{author}{Cazzaniga, P.}, \bibinfo{year}{2019}.
\newblock \bibinfo{title}{Medga: a novel evolutionary method for image
  enhancement in medical imaging systems}.
\newblock \bibinfo{journal}{Expert Systems with Applications}
  \bibinfo{volume}{119}, \bibinfo{pages}{387--399}.
\bibitem[{Schuler et~al.(2013)Schuler, Christopher~Burger, Harmeling and
  Scholkopf}]{schuler2013machine}
\bibinfo{author}{Schuler, C.J.}, \bibinfo{author}{Christopher~Burger, H.},
  \bibinfo{author}{Harmeling, S.}, \bibinfo{author}{Scholkopf, B.},
  \bibinfo{year}{2013}.
\newblock \bibinfo{title}{A machine learning approach for non-blind image
  deconvolution}, in: \bibinfo{booktitle}{Proceedings of the IEEE Conference on
  Computer Vision and Pattern Recognition}, pp. \bibinfo{pages}{1067--1074}.
\bibitem[{Schuler et~al.(2015)Schuler, Hirsch, Harmeling and
  Sch{\"o}lkopf}]{schuler2015learning}
\bibinfo{author}{Schuler, C.J.}, \bibinfo{author}{Hirsch, M.},
  \bibinfo{author}{Harmeling, S.}, \bibinfo{author}{Sch{\"o}lkopf, B.},
  \bibinfo{year}{2015}.
\newblock \bibinfo{title}{Learning to deblur}.
\newblock \bibinfo{journal}{IEEE transactions on pattern analysis and machine
  intelligence} \bibinfo{volume}{38}, \bibinfo{pages}{1439--1451}.
\bibitem[{Schwartz et~al.(2018)Schwartz, Giryes and
  Bronstein}]{schwartz2018deepisp}
\bibinfo{author}{Schwartz, E.}, \bibinfo{author}{Giryes, R.},
  \bibinfo{author}{Bronstein, A.M.}, \bibinfo{year}{2018}.
\newblock \bibinfo{title}{Deepisp: Toward learning an end-to-end image
  processing pipeline}.
\newblock \bibinfo{journal}{IEEE Transactions on Image Processing}
  \bibinfo{volume}{28}, \bibinfo{pages}{912--923}.
\bibitem[{Sha et~al.(2019)Sha, Zandavi and Chung}]{sha2019fast}
\bibinfo{author}{Sha, F.}, \bibinfo{author}{Zandavi, S.M.},
  \bibinfo{author}{Chung, Y.Y.}, \bibinfo{year}{2019}.
\newblock \bibinfo{title}{Fast deep parallel residual network for accurate
  super resolution image processing}.
\newblock \bibinfo{journal}{Expert Systems with Applications}
  \bibinfo{volume}{128}, \bibinfo{pages}{157--168}.
\bibitem[{Shan et~al.(2008)Shan, Jia and Agarwala}]{shan2008high}
\bibinfo{author}{Shan, Q.}, \bibinfo{author}{Jia, J.},
  \bibinfo{author}{Agarwala, A.}, \bibinfo{year}{2008}.
\newblock \bibinfo{title}{High-quality motion deblurring from a single image}.
\newblock \bibinfo{journal}{Acm transactions on graphics (tog)}
  \bibinfo{volume}{27}, \bibinfo{pages}{1--10}.
\bibitem[{Shao et~al.(2020)Shao, Liu, Ye, Wang, Ge, Bao and
  Li}]{shao2020deblurgan+}
\bibinfo{author}{Shao, W.Z.}, \bibinfo{author}{Liu, Y.Y.}, \bibinfo{author}{Ye,
  L.Y.}, \bibinfo{author}{Wang, L.Q.}, \bibinfo{author}{Ge, Q.},
  \bibinfo{author}{Bao, B.K.}, \bibinfo{author}{Li, H.B.},
  \bibinfo{year}{2020}.
\newblock \bibinfo{title}{Deblurgan+: Revisiting blind motion deblurring using
  conditional adversarial networks}.
\newblock \bibinfo{journal}{Signal Processing} \bibinfo{volume}{168},
  \bibinfo{pages}{107338}.
\bibitem[{Sharif et~al.(2021)Sharif, Naqvi and Biswas}]{sharif2021sagan}
\bibinfo{author}{Sharif, S.}, \bibinfo{author}{Naqvi, R.A.},
  \bibinfo{author}{Biswas, M.}, \bibinfo{year}{2021}.
\newblock \bibinfo{title}{Sagan: Adversarial spatial-asymmetric attention for
  noisy nona-bayer reconstruction}.
\newblock \bibinfo{journal}{arXiv preprint arXiv:2110.08619} .
\bibitem[{Sim and Kim(2019)}]{sim2019deep}
\bibinfo{author}{Sim, H.}, \bibinfo{author}{Kim, M.}, \bibinfo{year}{2019}.
\newblock \bibinfo{title}{A deep motion deblurring network based on per-pixel
  adaptive kernels with residual down-up and up-down modules}, in:
  \bibinfo{booktitle}{Proceedings of the IEEE Conference on Computer Vision and
  Pattern Recognition Workshops}, pp. \bibinfo{pages}{0--0}.
\bibitem[{Sun et~al.(2015)Sun, Cao, Xu and Ponce}]{sun2015learning}
\bibinfo{author}{Sun, J.}, \bibinfo{author}{Cao, W.}, \bibinfo{author}{Xu, Z.},
  \bibinfo{author}{Ponce, J.}, \bibinfo{year}{2015}.
\newblock \bibinfo{title}{Learning a convolutional neural network for
  non-uniform motion blur removal}, in: \bibinfo{booktitle}{Proceedings of the
  IEEE Conference on Computer Vision and Pattern Recognition}, pp.
  \bibinfo{pages}{769--777}.
\bibitem[{Szegedy et~al.(2016)Szegedy, Ioffe, Vanhoucke and
  Alemi}]{szegedy2016inception}
\bibinfo{author}{Szegedy, C.}, \bibinfo{author}{Ioffe, S.},
  \bibinfo{author}{Vanhoucke, V.}, \bibinfo{author}{Alemi, A.},
  \bibinfo{year}{2016}.
\newblock \bibinfo{title}{Inception-v4, inception-resnet and the impact of
  residual connections on learning}.
\newblock \bibinfo{journal}{arXiv preprint arXiv:1602.07261} .
\bibitem[{Tao et~al.(2018)Tao, Gao, Shen, Wang and Jia}]{tao2018scale}
\bibinfo{author}{Tao, X.}, \bibinfo{author}{Gao, H.}, \bibinfo{author}{Shen,
  X.}, \bibinfo{author}{Wang, J.}, \bibinfo{author}{Jia, J.},
  \bibinfo{year}{2018}.
\newblock \bibinfo{title}{Scale-recurrent network for deep image deblurring},
  in: \bibinfo{booktitle}{Proceedings of the IEEE Conference on Computer Vision
  and Pattern Recognition}, pp. \bibinfo{pages}{8174--8182}.
\bibitem[{Thakare et~al.(2018)Thakare, Kamble, Thengne and
  Kamble}]{thakare2018document}
\bibinfo{author}{Thakare, S.}, \bibinfo{author}{Kamble, A.},
  \bibinfo{author}{Thengne, V.}, \bibinfo{author}{Kamble, U.},
  \bibinfo{year}{2018}.
\newblock \bibinfo{title}{Document segmentation and language translation using
  tesseract-ocr}, in: \bibinfo{booktitle}{2018 IEEE 13th International
  Conference on Industrial and Information Systems (ICIIS)},
  \bibinfo{organization}{IEEE}. pp. \bibinfo{pages}{148--151}.
\bibitem[{Vedaldi and Fulkerson(2010)}]{vedaldi2010vlfeat}
\bibinfo{author}{Vedaldi, A.}, \bibinfo{author}{Fulkerson, B.},
  \bibinfo{year}{2010}.
\newblock \bibinfo{title}{Vlfeat: An open and portable library of computer
  vision algorithms}, in: \bibinfo{booktitle}{Proceedings of the 18th ACM
  international conference on Multimedia}, pp. \bibinfo{pages}{1469--1472}.
\bibitem[{Wang et~al.(2012)Wang, Lu, Wang and Jia}]{wang2012kernel}
\bibinfo{author}{Wang, J.}, \bibinfo{author}{Lu, K.}, \bibinfo{author}{Wang,
  Q.}, \bibinfo{author}{Jia, J.}, \bibinfo{year}{2012}.
\newblock \bibinfo{title}{Kernel optimization for blind motion deblurring with
  image edge prior}.
\newblock \bibinfo{journal}{Mathematical Problems in Engineering}
  \bibinfo{volume}{2012}.
\bibitem[{Wang and Shen(2017)}]{wang2017deep}
\bibinfo{author}{Wang, W.}, \bibinfo{author}{Shen, J.}, \bibinfo{year}{2017}.
\newblock \bibinfo{title}{Deep visual attention prediction}.
\newblock \bibinfo{journal}{IEEE Transactions on Image Processing}
  \bibinfo{volume}{27}, \bibinfo{pages}{2368--2378}.
\bibitem[{Wang et~al.(2018)Wang, Yu, Wu, Gu, Liu, Dong, Qiao and
  Change~Loy}]{wang2018esrgan}
\bibinfo{author}{Wang, X.}, \bibinfo{author}{Yu, K.}, \bibinfo{author}{Wu, S.},
  \bibinfo{author}{Gu, J.}, \bibinfo{author}{Liu, Y.}, \bibinfo{author}{Dong,
  C.}, \bibinfo{author}{Qiao, Y.}, \bibinfo{author}{Change~Loy, C.},
  \bibinfo{year}{2018}.
\newblock \bibinfo{title}{Esrgan: Enhanced super-resolution generative
  adversarial networks}, in: \bibinfo{booktitle}{Proceedings of the European
  Conference on Computer Vision (ECCV)}, pp. \bibinfo{pages}{0--0}.
\bibitem[{Wei et~al.(2018)Wei, Wang, Yang and Liu}]{wei2018deep}
\bibinfo{author}{Wei, C.}, \bibinfo{author}{Wang, W.}, \bibinfo{author}{Yang,
  W.}, \bibinfo{author}{Liu, J.}, \bibinfo{year}{2018}.
\newblock \bibinfo{title}{Deep retinex decomposition for low-light
  enhancement}.
\newblock \bibinfo{journal}{arXiv preprint arXiv:1808.04560} .
\bibitem[{Whyte et~al.(2012)Whyte, Sivic, Zisserman and Ponce}]{whyte2012non}
\bibinfo{author}{Whyte, O.}, \bibinfo{author}{Sivic, J.},
  \bibinfo{author}{Zisserman, A.}, \bibinfo{author}{Ponce, J.},
  \bibinfo{year}{2012}.
\newblock \bibinfo{title}{Non-uniform deblurring for shaken images}.
\newblock \bibinfo{journal}{International journal of computer vision}
  \bibinfo{volume}{98}, \bibinfo{pages}{168--186}.
\bibitem[{Xu and Jia(2010)}]{xu2010two}
\bibinfo{author}{Xu, L.}, \bibinfo{author}{Jia, J.}, \bibinfo{year}{2010}.
\newblock \bibinfo{title}{Two-phase kernel estimation for robust motion
  deblurring}, in: \bibinfo{booktitle}{European conference on computer vision},
  \bibinfo{organization}{Springer}. pp. \bibinfo{pages}{157--170}.
\bibitem[{Xu et~al.(2012)Xu, Wang, Hu and Peng}]{xu2012single}
\bibinfo{author}{Xu, Y.}, \bibinfo{author}{Wang, L.}, \bibinfo{author}{Hu, X.},
  \bibinfo{author}{Peng, S.}, \bibinfo{year}{2012}.
\newblock \bibinfo{title}{Single-image blind deblurring for non-uniform
  camera-shake blur}, in: \bibinfo{booktitle}{Asian Conference on Computer
  Vision}, \bibinfo{organization}{Springer}. pp. \bibinfo{pages}{336--348}.
\bibitem[{Xu et~al.(2017)Xu, Ye, Cui, Feng, Li and Chen}]{xu2017image}
\bibinfo{author}{Xu, Z.}, \bibinfo{author}{Ye, P.}, \bibinfo{author}{Cui, G.},
  \bibinfo{author}{Feng, H.}, \bibinfo{author}{Li, Q.}, \bibinfo{author}{Chen,
  Y.}, \bibinfo{year}{2017}.
\newblock \bibinfo{title}{Image restoration for large-motion blurred lunar
  remote sensing image}, in: \bibinfo{booktitle}{Selected Papers of the Chinese
  Society for Optical Engineering Conferences held October and November 2016},
  \bibinfo{organization}{International Society for Optics and Photonics}. p.
  \bibinfo{pages}{1025534}.
\bibitem[{Ye et~al.(2020)Ye, Lyu and Chen}]{ye2020scale}
\bibinfo{author}{Ye, M.}, \bibinfo{author}{Lyu, D.}, \bibinfo{author}{Chen,
  G.}, \bibinfo{year}{2020}.
\newblock \bibinfo{title}{Scale-iterative upscaling network for image
  deblurring}.
\newblock \bibinfo{journal}{IEEE Access} \bibinfo{volume}{8},
  \bibinfo{pages}{18316--18325}.
\bibitem[{Yu et~al.(2019)Yu, Lin, Yang, Shen, Lu and Huang}]{yu2019free}
\bibinfo{author}{Yu, J.}, \bibinfo{author}{Lin, Z.}, \bibinfo{author}{Yang,
  J.}, \bibinfo{author}{Shen, X.}, \bibinfo{author}{Lu, X.},
  \bibinfo{author}{Huang, T.S.}, \bibinfo{year}{2019}.
\newblock \bibinfo{title}{Free-form image inpainting with gated convolution},
  in: \bibinfo{booktitle}{Proceedings of the IEEE International Conference on
  Computer Vision}, pp. \bibinfo{pages}{4471--4480}.
\bibitem[{Zhang et~al.(2018a)Zhang, Pan, Ren, Song, Bao, Lau and
  Yang}]{zhang2018dynamic}
\bibinfo{author}{Zhang, J.}, \bibinfo{author}{Pan, J.}, \bibinfo{author}{Ren,
  J.}, \bibinfo{author}{Song, Y.}, \bibinfo{author}{Bao, L.},
  \bibinfo{author}{Lau, R.W.}, \bibinfo{author}{Yang, M.H.},
  \bibinfo{year}{2018}a.
\newblock \bibinfo{title}{Dynamic scene deblurring using spatially variant
  recurrent neural networks}, in: \bibinfo{booktitle}{Proceedings of the IEEE
  Conference on Computer Vision and Pattern Recognition}, pp.
  \bibinfo{pages}{2521--2529}.
\bibitem[{Zhang et~al.(2018b)Zhang, Tian, Kong, Zhong and
  Fu}]{zhang2018residual}
\bibinfo{author}{Zhang, Y.}, \bibinfo{author}{Tian, Y.}, \bibinfo{author}{Kong,
  Y.}, \bibinfo{author}{Zhong, B.}, \bibinfo{author}{Fu, Y.},
  \bibinfo{year}{2018}b.
\newblock \bibinfo{title}{Residual dense network for image super-resolution},
  in: \bibinfo{booktitle}{Proceedings of the IEEE conference on computer vision
  and pattern recognition}, pp. \bibinfo{pages}{2472--2481}.
\bibitem[{Zhao et~al.(2016)Zhao, Gallo, Frosio and Kautz}]{zhao2016loss}
\bibinfo{author}{Zhao, H.}, \bibinfo{author}{Gallo, O.},
  \bibinfo{author}{Frosio, I.}, \bibinfo{author}{Kautz, J.},
  \bibinfo{year}{2016}.
\newblock \bibinfo{title}{Loss functions for image restoration with neural
  networks}.
\newblock \bibinfo{journal}{IEEE Transactions on computational imaging}
  \bibinfo{volume}{3}, \bibinfo{pages}{47--57}.

\end{thebibliography}

\end{document}